\pdfoutput=1

\documentclass[11pt]{article}

\usepackage[preprint]{acl}

\usepackage{times}
\usepackage{pgfplots}
\usepackage{latexsym}
\usepackage{amssymb}
\usepackage{amsmath}
\usepackage{pifont}
\usepackage[T1]{fontenc}
\usepackage[utf8]{inputenc}
\usepackage{microtype}
\usepackage{inconsolata}
\usepackage{graphicx}

\usepackage[inline]{enumitem}
\usepackage{booktabs}
\usepackage{makecell}
\usepackage{standalone}
\usepackage{caption}
\usepackage{xcolor}
\usepackage{xspace}
\usepackage{multirow}

\usepackage[most]{tcolorbox}
\usepackage{etoolbox}
\tcbuselibrary{breakable}

\pgfplotsset{compat=1.18}
\usepgfplotslibrary{statistics, groupplots}
\usepgflibrary{plotmarks}
\usetikzlibrary{calc, patterns, matrix, positioning, shapes, shadows, shapes.geometric, arrows.meta}

\definecolor{ibm1}{HTML}{0077BB}
\definecolor{ibm2}{HTML}{33BBEE}
\definecolor{ibm3}{HTML}{EE7733}
\definecolor{ibm4}{HTML}{EE3377}
\definecolor{ibm5}{HTML}{CC3311}

\definecolor{grad1}{HTML}{364B9A}
\definecolor{grad2}{HTML}{6EA6CD}
\definecolor{grad3}{HTML}{C2E4EF}
\definecolor{grad4}{HTML}{FEDA8B}
\definecolor{grad5}{HTML}{F67E4B}

\newtcolorbox[use counter=prompt]{promptbox}[2][]%
   {
    enhanced jigsaw,
    breakable,
    boxrule=0mm,
    colback=white, colframe=gray!50!black,
    top=0mm,bottom=1mm,left=1mm,right=1mm,
    arc=1mm,
    title={\textbf{Prompt \thetcbcounter: #2}},
    borderline={0.5pt}{0pt}{gray!50!black, rounded corners},
    #1,
}
\AtBeginEnvironment{promptbox}{\small \sffamily}

\title{\textsc{HintsOfTruth}: A Multimodal Checkworthiness Detection Dataset with Real and Synthetic Claims}

\newcommand{\university}{\includegraphics[height=1.4em]{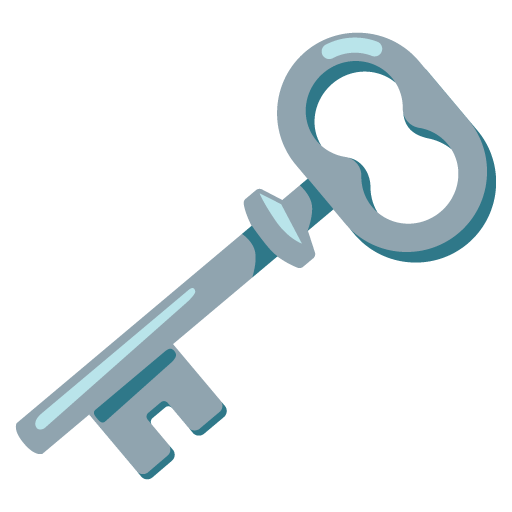}}
\newcommand{\institute}{\includegraphics[height=1.4em]{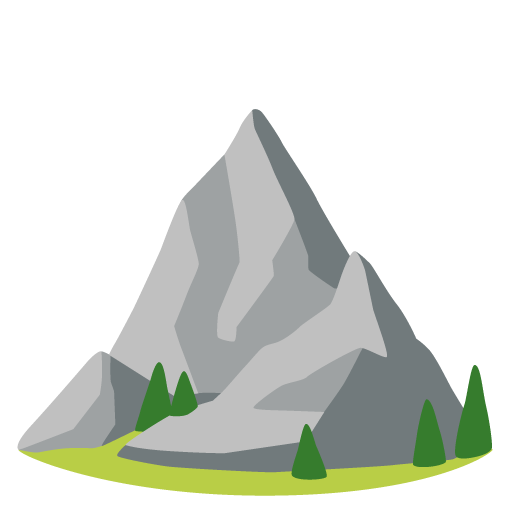}}
\newcommand{\lugano}{\includegraphics[height=1.4em]{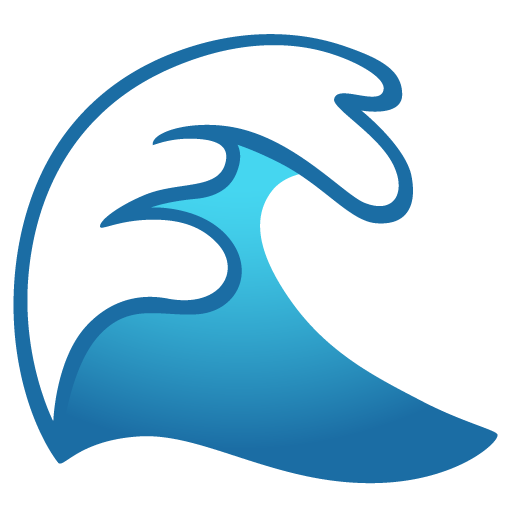}}

\author{
 \textbf{Michiel van der Meer\textsuperscript{\university,\institute}},
 \textbf{Pavel Korshunov\textsuperscript{\institute}}, \\
 \textbf{S\'{e}bastien Marcel\textsuperscript{\institute}},
 \textbf{Lonneke van der Plas\textsuperscript{\lugano,\institute}}
\\
\\
 {\small \institute} Idiap Research Institute, Martigny, Switzerland,\\
 {\small \university} Leiden University, Leiden, The Netherlands,\\
 {\small \lugano} USI Università della Svizzera italiana, Lugano, Switzerland
}

\newcommand{\cmark}{\ding{51}}%
\newcommand{\xmark}{\ding{55}}%

\newcommand{\hiot}{\textsc{HintsOfTruth}\xspace}

\begin{document}
\maketitle
\begin{abstract}
Misinformation can be countered with fact-checking, but the process is costly and slow. Identifying checkworthy claims is the first step, where automation can help scale fact-checkers' efforts. However, detection methods struggle with content that is
\begin{enumerate*}[label=(\arabic*)]
    \item multimodal,
    \item from diverse domains, and
    \item synthetic
\end{enumerate*}.
We introduce \hiot, a public dataset for multimodal checkworthiness detection with $27$K real-world and synthetic image/claim pairs. The mix of real and synthetic data makes this dataset unique and ideal for benchmarking detection methods.
We compare fine-tuned and prompted Large Language Models (LLMs). We find that well-configured lightweight text-based encoders perform comparably to multimodal models but the former only focus on identifying non-claim-like content. Multimodal LLMs can be more accurate but come at a significant computational cost, making them impractical for large-scale applications. When faced with synthetic data, multimodal models perform more robustly. \\[1mm]
\includegraphics[width=.9em,trim=0 {1cm} 0 0]{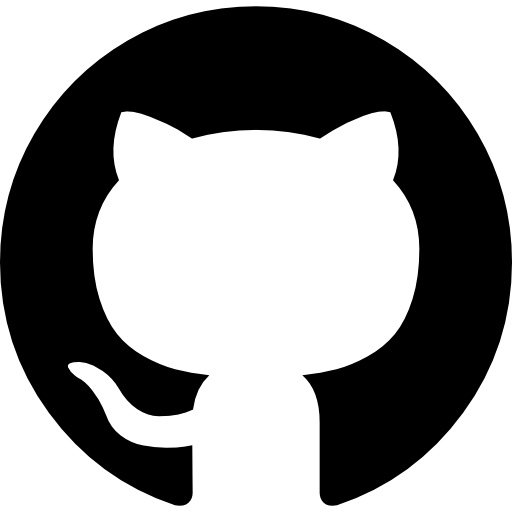} \url{https://hintsoftruth.github.io/}
\end{abstract}

\section{Introduction}
\label{sec:intro}
Online misinformation spreads rapidly via social networks and deceptive websites posing as legitimate news sources \citep{del2016spreading, rocha2021impact, Ecker2024}. This influences voting behavior \citep{ribeiro2017everything} and pollutes the digital information space \citep{greenspan2021pandemics, sharma2019combating}. Misinformation tactics include decontextualization (e.g., wrongly presenting image-based evidence) and providing incomplete information \citep{kreps2022all}. Generative AI, like ChatGPT \citep{openai2023chatgpt} for text and Midjourney \citep{midjourney2023midjourney} for images, has worsened the issue by enabling large-scale alteration or fabrication of news narratives \citep{zhou2023synthetic, chen2024combating}. Given these developments, continuous verification of multimodal information is a key challenge \citep{abdelnabi2022open, singh2022predicting}.

\begin{figure}[t]
    \centering
    \begin{tikzpicture}
        \node[] at (0,0) {\includegraphics[width=4cm]{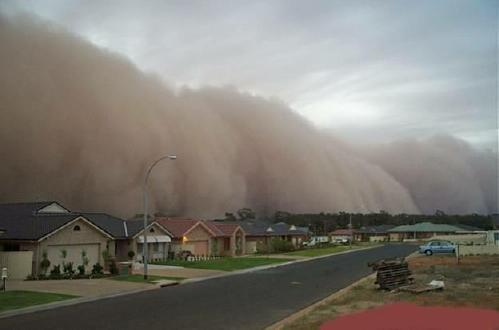}};
        \node[draw=black, align=center, text width=3cm] at (4,0) {photograph shows a tsunami 1/2 second before it struck the island of Sumatra.};
        \node[] at (2, 1.5) {\textbf{Checkworthy \cmark}};

        \node[] at (0,-3.4) {\includegraphics[width=4cm]{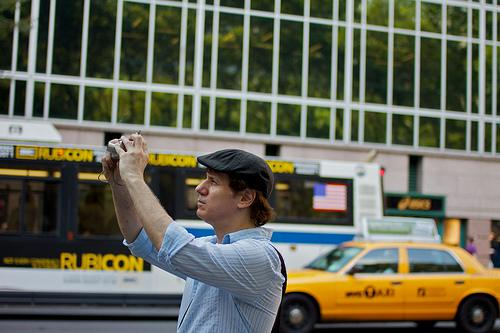}};
        \node[draw=black, align=center, text width=3cm] at (4,-3.4) {a man in a black hat and blue shirt taking a picture on the side of an urban street.};
        \node[] at (2, -1.8) {\textbf{Non-checkworthy \xmark}};
    \end{tikzpicture}
    \caption{Examples of the multimodal checkworthiness detection task.}
    \label{fig:example-first-page}
\end{figure}

Media gatekeepers, including news publishers and fact-checking services, verify content veracity, but manual fact-checking is costly and time-consuming \citep{nakov2021automated}. Therefore, selecting which claims to fact-check is a major challenge, as the amount of potential misinformation far exceeds fact-checking capacity. Automated approaches can help by identifying \textbf{checkworthy} claims \citep{nakov2018overview, konstantinovskiy2021toward}, see Figure~\ref{fig:example-first-page} for an example, or in other stages in the fact-checking pipeline (Figure~\ref{fig:fact-checking-pipeline}).

However, existing automated checkworthiness detection methods \begin{enumerate*}[label=(\arabic*)]
    \item have poor support for multimodal content,
    \item have only been tested in a limited number of domains,
    \item have unknown capabilities on synthetic media, and
    \item do not consider compute cost as a factor.
\end{enumerate*} \citep{akhtar-etal-2023-multimodal}.
First, modern misinformation often includes mixed forms of media, such as images or videos \citep{dufour2024ammeba}, yet it is unclear if detection methods effectively integrate visual data \citep{alam2023overview}.
Second, strategies for misinformation detection vary by domain \citep{ecker2022psychological, chen2021persuasion, lasser2023alternative}, raising concerns about generalizability, especially, for practical applications \citep{jiang2018linguistic, monteith2024artificial}.
Third, ubiquitous access to generative models is reshaping misinformation \citep{xu2023combating}, warranting the evaluation of detection methods on synthetic content.
Lastly, while Large Language Models (LLMs) perform well, their high compute cost may render large-scale checkworthiness detection impractical \citep{augenstein2024factuality}, though the exact tradeoffs are unknown.

This paper introduces \hiot, the first publicly available multimodal dataset of image-text pairs containing both real-world and synthetically generated checkworthy and non-checkworthy claims. We source real claims from datasets like 5Pils \citep{tonglet2024image}, Multiclaim \citep{pikuliak2023multilingual}, Flickr30K \citep{hodosh2013framing}, and SentiCap \citep{sharma2018conceptual}. Synthetic images and text are generated using Flux \citep{flux2024}, StableDiffusion 3.5 \citep[][SD]{stablediff}, Llava \citep{li2024llavanext}, and BLIP \citep{li2022blip}. We evaluate recent text and image models, from lightweight ones like TinyBERT \citep{tinybert} for scalability to large, multimodal models like Pixtral \citep{mistral2024pixtral}. These evaluations reveal model limitations and guide practical decisions in checkworthiness detection.

\paragraph{Contributions}
We present: \begin{enumerate*}[label=(\arabic*)]
    \item \hiot, a novel dataset for multimodal checkworthiness detection from diverse sources, with an established connection between images and textual claims, that can be used as a benchmark for checkworthiness detection models,
    \item synthetic counterparts of images and claims in the dataset, which has not been explored in the context of checkworthiness, and
    \item an extensive set of experiments demonstrating the limits of state-of-the-art detection methods.
\end{enumerate*}

\section{Related Work}
\subsection{Human-Centered Fact-Checking}
Recent research on human-centered AI has emphasized developing tools that augment humans \citep{akata2020research, nakov2021automated}. In the field of fact-checking, such tools would complement human fact-checkers in their work \citep{micallef2022true, graves2017anatomy}, allowing experts to control \emph{what} and \emph{how} to fact-check \citep{das2023state}.

\begin{figure}
    \centering
\resizebox{\columnwidth}{!}{%
\begin{tikzpicture}[node distance=1cm and 0.5cm, auto]
    \tikzstyle{startstop} = [
        rectangle,
        rounded corners,
        text width=2cm,
        minimum height=1cm,
        text centered,
        draw=ibm1!60,
        fill=ibm1!10,
        font=\sffamily,
        drop shadow
    ]
    \tikzstyle{expert-inv} = [
        rectangle,
        minimum width=5cm,
        minimum height=1cm,
        draw=ibm5!60,
        fill=ibm5!30,
    ]
    \tikzstyle{process} = [
        rectangle,
        text width=1.6cm,
        minimum height=1cm,
        text centered,
        draw=ibm2!60,
        fill=ibm2!30,
        font=\sffamily,
        drop shadow
    ]
    \tikzstyle{text-only} = [
        font=\sffamily,
    ]
    \tikzstyle{highlight} = [
        rectangle,
        minimum width=2.3cm,
        minimum height=2.3cm,
        draw=ibm1,
        rounded corners,
        line width=3pt,
        fill=none,
    ]

    \tikzstyle{arrow} = [line width=1.3pt, ->, >=Stealth, draw=black!80]

    \node (input) [startstop] {\small \textbf{Input}\\  (Text, Image, Video, Audio)};
    \node (stage1) [process, right=of input] {\small Claim \& Checkworthiness Detection};
    \node (stage2) [process, right=of stage1] {\small Evidence Retrieval};
    \node (stage3) [process, right=of stage2] {\small Verdict Prediction\\ \& Justification};
    \node (highlight) [highlight, dashed, dash pattern=on 5pt off 3pt, below right=0cm and 0cm of stage1.base, xshift=-1.2cm, yshift=.55cm] {};

    \draw [arrow] (input) -- (stage1);
    \draw [arrow] (stage1) -- (stage2);
    \draw [arrow] (stage2) -- (stage3);

    \node (data-top-left) [above=2.5cm of input, xshift=1.5cm] {};
    \node (data-top-right) [right=6.5cm of data-top-left] {};
    \node (data-bottom-left) [below=1.66cm of data-top-left] {};
    \fill[ibm3, rounded corners=10pt, opacity=0.3]
        (data-top-left.center) --
        (data-top-right.center) --
        (data-bottom-left.center) --
        cycle;
    \node (exp-bottom-left) [below=.001cm of data-bottom-left] {};
    \node (exp-top-right) [below=.001cm of data-top-right] {};
    \node (exp-bottom-right) [below=1.66cm of exp-top-right] {};
    \fill[ibm4, rounded corners=10pt, opacity=0.3]
        (exp-bottom-left.center) --
        (exp-top-right.center) --
        (exp-bottom-right.center) --
        cycle;
    \node [text-only, below right=.05cm and .1cm of data-top-left] {Data involvement};
    \node [text-only, above left=.05cm and .1cm of exp-bottom-right] {Expert involvement};
\end{tikzpicture}%
}
    \caption{The fact-checking pipeline from \citet{akhtar-etal-2023-multimodal}, visualized the amount of data and expert effort required. We focus on the highlighted stage. }
    \label{fig:fact-checking-pipeline}
\end{figure}
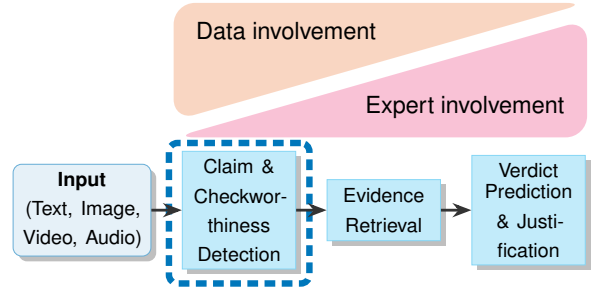


Crucially, as shown in Figure~\ref{fig:fact-checking-pipeline}, the fact-checking pipeline involves handling large amounts of data and needs expert involvement. In the early stages of the pipeline, large amounts of data are processed to filter out \textbf{checkworthy} content. Hence, relying on experts to do this task manually is infeasible. Moreover, the expert is required in the later stages for the complex tasks of verdict prediction and justification. Natural Language Processing (NLP) technology provides various types of support, especially when dealing with scale \citep{vandermeer2024facilitating, procter2023some}, to simplify the problem \citep{chen2022generating, bonet2024run}, or to combat cognitive biases \citep{soprano2024cognitive}. In this work, we use NLP techniques to address the scale issues for checkworthiness detection.


\subsection{Misinformation in the Age of LLMs}
LLMs play a significant role both in detecting and generating misinformation. Recent work integrates LLMs into fact-checking frameworks \citep{geng2024multimodal}, although methods are shown to generalize poorly across time \citep{stepanova-ross-2023-temporal}. Nonetheless, LLMs look promising when applied to text-based checkworthiness detection \citep{majer2024claim}. Reviews of LLM-generated multimedia highlight the open challenges \citep{lin2024detecting, augenstein2024factuality}. For instance, large amounts of synthetic misinformation have the potential to impact the quality of future LLMs \citep{pan2023risk}, and misinformation generated by GPT-4 may be harder to detect than that written by humans \citep{chen2024combating}.

\subsection{Multimodal Resources}
Existing work on fact-checking emphasizes empirical research, which involves extensively benchmarking fact-checking methods \citep{schlichtkrull2024averitec, papadopoulos2024verite}, often using distant supervision \citep{nakamura2020fakeddit, zlatkova2019fact}. Most multimodal datasets investigate the out-of-context use of images and claims \citep{luo2021newsclippings, tonglet2024image}, or whether claims are reflected in an image \citep{yoon2024assessing, papadopoulos2023synthetic}. Few datasets exist that
\begin{enumerate*}[label=(\arabic*)]
    \item check whether the image contributes new information \citep{liu2024mmfakebench}, or
    \item contain synthetically generated data \citep{xu2023combating, seow2022comprehensive}
\end{enumerate*}. The few efforts on multimodal checkworthiness indicate that textual data, whether through OCR or by focusing on claims only, is sufficient for state-of-the-art performance \citep{frick2023fraunhofer}. More extensive experiments on varied types of data with complex image use, across domains, are needed to further examine this finding.

\section{Method}
\label{sec:method}
We introduce the multimodal checkworthiness task definition, how we obtain the real-world data underlying \hiot, and how we generate synthetic samples to augment our dataset.

\subsection{Task Definition: Multimodal Checkworthiness Detection}
\label{sec:task-definition}
Given a textual claim $c$ and an image $i$ published alongside the claim, our task aims to predict whether the pair is worthy of fact-checking $p(i,c) = 1$. In checkworthiness detection, the following questions are answered:
\begin{enumerate*}[label=(\textbf{Q\arabic*})]
    \item Does the text contain a verifiable factual claim?
    \item Is the claim potentially harmful, urgent, and up-to-date?
\end{enumerate*}
The task definition is derived from \citet{barron2020overview}, which also formed the basis for the canonical dataset for multimodal checkworthiness detection, CheckThat! 2023 Task 1A \citep{alam2023overview, cheema2022mm}.

To establish that an image provides meaningful context to a claim and is necessary for assessing the pair's checkworthiness, we also consider the following contextualized questions:
\begin{enumerate*}[label=(\textbf{Q\arabic*})]
    \setcounter{enumi}{2}
    \item Is the content of the claim reflected in the image?
    \item Does the image contribute extra information to the claim?
\end{enumerate*}
These two questions help identify \emph{complex} image use, which will test the multimodal capabilities of checkworthiness detectors \citep{dufour2024ammeba}.

\subsection{Getting Checkworthy Image/Claim Pairs}
We set out to obtain image/claims pairs that we deem checkworthy. We rely on data stemming from fact-checking articles, as claims in these articles have already been checked. Fact-checking articles are written by experienced fact-checkers and contain rich contextual information. In practice, claims are often sourced from social media platforms. We obtain our data from two sources:
\begin{description}[itemsep=-5pt, leftmargin=0pt, topsep=0pt, partopsep=0pt]
    \item[5Pils] \citep{tonglet2024image}. 5Pils contains extracted images, claims, and contextual questions about claims from news sources in India, Kenya, and South Sudan. Through the use of contextual questions, images in this dataset are ensured to have a relationship with the claim.
    \item[Multiclaim] \citep{pikuliak2023multilingual} contains URLs to a wide array of fact-checking articles and their respective claims but needs to be scraped and filtered for images. We retain those claims for which \begin{enumerate*}[label=(\arabic*)]
    \item we find images in close proximity, and
    \item explicitly refer to visual information
    \end{enumerate*}. See Appendix~\ref{app:multiclaim-extraction} for additional details.
\end{description}

\subsection{Non-checkworthy Image/Claim Pairs}
We also need image/text pairs that are not checkworthy. We resort to strategies derived from the task definition for obtaining negative instances. We select samples from datasets that we consider not checkworthy because they answer `no' to any of the guiding questions posed in Section~\ref{sec:task-definition}. The strategies select:
\begin{description}[itemsep=-5pt, leftmargin=0pt, topsep=0pt, partopsep=0pt]
    \item[Non-factual (Q1)] claims, such as subjective opinions or facts that cannot be verified using external information. The dataset representing this strategy is \textbf{SentiCap} \citep{sharma2018conceptual}.
    \item[Non-relevant (Q2)] statements that are not harmful, not about breaking news, not up-to-date, or not relevant to news topics. The dataset representing this strategy is \textbf{Flickr30K} \citep{hodosh2013framing}, though there are many other resources containing arbitrary image-text pairs (see Appendix~\ref{app:image-captions}).
    \item[No cross-modal connection (Q3)] images we know have a deep connection with a text but with the image swapped to no longer make sense. The dataset representing this strategy is \textbf{Fakeddit} \citep{nakamura2020fakeddit}.
\end{description}
To incorporate the fourth guiding question (\textbf{Q4}), we filter out claims from any of the stated datasets that do not explicitly refer to multimodal content (see Appendix~\ref{app:multimodal-terms} for a list of terms). This way, we encourage that the samples with basic image use (i.e., those pairs where the claim does not refer to the image) are excluded from our dataset. We combine the checkworthy and non-checkworthy samples into \hiot, our novel multimodal checkworthiness dataset that spans multiple domains.

\begin{table*}[ht]
  \centering
  \resizebox{\textwidth}{!}{%
  \begin{tabular}{@{}l l c r p{7.7cm}@{}}
    \toprule
    \textbf{Dataset} & \textbf{Source / Subset} & \textbf{Checkworthy} & \textbf{Size} & \textbf{Description} \\
    \midrule
    CheckThat 2023 Task 1A
      & Twitter
      & Mixed
      & 3,175
      & Tweets on COVID-19, technology, climate change. \\[1mm]
    \midrule
      & 5Pils
      & \cmark
      & 1,676
      & News articles from India, Kenya, and South Sudan. \\[0.5mm]
      & Multiclaim
      & \cmark
      & 3,048
      & Social media posts in a general domain. \\[0.5mm]
      & SentiCap
      & \xmark
      & 3,171
      & Captions with sentiment injection. \\[0.5mm]
      & Flickr30K
      & \xmark
      & 3,000
      & Image captions from a general domain. \\[0.5mm]
      & Fakeddit
      & \xmark
      & 1,382
      & Reddit posts from a general domain. \\[1mm]
      \cmidrule{2-5}
      \hiot & Mixed & Mixed & 12,277 & Mixed domain benchmark\\
    \midrule
    \multirow{2}{*}{\(\hiot\text{-aug}\)}
      & 5Pils
      & Mixed
      & 1,676
      & Generated claims using BLIP, Llava. Generated images using Flux, StableDiffusion 3.5. \\[0.5mm]
      & Flickr30K
      & \xmark
      & 3,000
      & Generated captions using BLIP, Llava. Generated images using Flux, StableDiffusion 3.5. \\
    \bottomrule
  \end{tabular}%
  }
  \caption{Overview of the datasets used in this study for multimodal checkworthiness. \(\hiot\) aggregates samples from five sources, and \(\hiot\text{-aug}\) contains synthetically generated variants. }
  \label{tab:datasets}
\end{table*}

\subsection{Generating Synthetic Samples}
\label{sec:method-generating}
Given the risks of synthetic misinformation \citep{dufour2024ammeba, papadopoulos2023synthetic, zhou2023synthetic}, we augment our dataset with additional samples that contain either claims or images generated using various publicly accessible models. New samples consist of the original text (or image) and the corresponding generated image (or text).
Specifically, we employ two image generators to create images from claims and two multimodal models to generate claims from images. Our approach follows a simple cross-modal generation method: models freely generate corresponding text or images without requiring adversarial prompts \citep{perez2022ignore}. This allows us to examine how checkworthiness detection models respond to synthetic data. The labels of the new samples depend solely on the claim: synthetic \textbf{claims} are deemed \emph{non-checkworthy}, as models primarily generate non-relevant captions (see \textbf{Q2} in Section~\ref{sec:task-definition}), while synthetic \textbf{images} retain their original label, ensuring consistency with the claim’s content.








\section{Experiments}

\subsection{Data}
See Table~\ref{tab:datasets} for the datasets used in this paper. We use the canonical CheckThat! 2023 Task 1A dataset \citep[`CheckThat' henceforth,][]{alam2023overview} as training dataset and reference benchmark using its predefined train/test split, which represents the in-distribution scenario (models are trained and tested on the same dataset). CheckThat has a label ratio of .66/.34 between non-checkworthy and checkworthy samples. The \hiot dataset is split into two equally-sized development sets, and a smaller test set (40\%, 40\%, and 20\%, respectively). We use the test set (label ratio of .62/.38) for testing the detection methods fine-tuned on CheckThat. This represents a cross-distribution scenario, which should be more challenging than an in-distribution setup. \hiot is available online\footnote{\url{https://huggingface.co/datasets/michiel/hints_of_truth}}.

\subsection{Multimodal Checkworthiness Methods}
\label{sec:experimental-approaches}
We experiment with various state-of-the-art text-based, image-based, and multimodal encoders for checkworthiness detection, see Table~\ref{tab:models}. We use different model sizes to investigate the tradeoff between compute cost and task performance. We include single-modality models to identify whether both modalities are needed (i.e., checkworthiness can be assessed without leveraging cross-modal information). In addition, we distinguish between encoder-only and decoder-only models, to determine the difference between fine-tuning models on multimodal checkworthiness and In-Context Learning \citep[][]{dong2022survey}. Below, we describe the experimental setup for each type of approach. Additional information is available in Appendix~\ref{app:experimental-details}.
\begin{table}[t]
    \centering
    \begin{tabular}{@{}lllc@{}}
        \toprule
        \textbf{Model} & \textbf{Modality} & \textbf{Size} & \textbf{App.} \\
        \midrule
        TinyBERT & text & 14M & FT\\
        BERT-base & text & 109M & FT\\
        BERT-large & text & 335M & FT\\
        \midrule
        ResNet-26 & image & 16M & FT \\
        ViT-base & image & 86M & FT\\
        ViT-large & image & 303M & FT\\
        \midrule
        BLIP & text, image & 385M & FT\\
        BLIP2 & text, image & 1.17B & FT\\
        \midrule
        Llava  & text, image & 7.57B & ICL\\
        Pixtral & text, image & 12.4B & ICL\\
        \bottomrule
    \end{tabular}
    \caption{Models used for checkworthiness detection. Depending on the model, we fine-tune them (FT) or perform In-Context Learning (ICL).}
    \label{tab:models}
\end{table}

\paragraph{Fine-Tuning (FT)}
To fine-tune models for multimodal checkworthiness detection, we update all model parameters $\theta$ when predicting $p_{\theta}(i,c)$. We instantiate the models using pretrained versions, adding a single linear classification layer with a two-node output over their embeddings.\footnote{Single-logit was less stable and yielded inferior performance, see App. \ref{app:single-logit}.} We fine-tune our models with data from CheckThat using its predefined train/val/test split. Additionally, we tune a threshold parameter on the positive class probability, similar to a single-neuron sigmoid output \citep{ZOU20162, Korshunov_ICB_2019}. At various thresholds, we compute the True Positive Rate (TPR) and False Positive Rate (FPR), and we select the threshold for an FPR of $0.3$, prioritizing recall over precision (see App.~\ref{app:tuning-threshold}).

\paragraph{In-Context Learning (ICL)}
We evaluate the impact of $n$-shot learning (with $n=\{0,1,2,5\}$) and prompt verbosity. The verbose prompt instructions include guiding questions \textbf{Q1} through \textbf{Q4} (see Section~\ref{sec:task-definition}), while the succinct prompt only asks for an overall checkworthiness label. We experiment with two models:
\begin{enumerate*}[label=(\arabic*)]
    \item \textbf{Llava} \citep{liu2024visual}, using a Mistral-7B backend with a 32K token context.
    \item \textbf{Pixtral} \citep{mistral2024pixtral}, with a context size of 1024K.
\end{enumerate*}
Both models are chosen for their compatibility with standard hardware (up to a single H100 with 80GB VRAM) and accessibility, excluding non-local proprietary LLMs. Our experiments focus on zero-shot ICL with concise instructions to minimize token usage, though multiple setups are explored in Section~\ref{sec:cross-modal}.

\subsection{Research Questions}
Based on the criteria discussed in Section~\ref{sec:intro}, we conduct four experiments to address:
\begin{enumerate*}[label=(RQ\arabic*)]
\item Does combining modalities influence checkworthiness detection performance?
\item How well do models generalize across domains?
\item How do models fare on synthetic data?
\item What is the tradeoff between compute cost and task performance?
\end{enumerate*}

\section{Results}
Table~\ref{tab:checkthat20231a} shows the in-distribution experiments results on CheckThat, and Table~\ref{tab:cross-dataset-perf} demonstrates the results on the non-synthetic, real part of \hiot, illustrating the cross-distribution experiments. We answer each research question individually step by step.

\begin{table}
    \centering
    \resizebox{\columnwidth}{!}{%
    \begin{tabular}{@{}lcccc@{}}
        \toprule
        \textbf{Model} & \textbf{Prec.} & \textbf{Rec.} & \textbf{F1} & \textbf{Acc.}\\
        \midrule
        TinyBERT         & 0.698 & 0.721 & 0.702 & 0.724\\
        BERT-base        & \underline{0.735} & 0.769 & 0.735 & 0.748\\
        BERT-large       & 0.726 & 0.760 & 0.723 & 0.735\\
        ResNet           & 0.595 & 0.600 & 0.596 & 0.641\\
        ViT-base         & 0.639 & 0.655 & 0.640 & 0.666\\
        ViT-large        & 0.654 & 0.670 & 0.658 & 0.686\\
        BLIP             &\textbf{0.782} & \underline{0.819} & \textbf{0.788} & \textbf{0.801}\\
        BLIP2            & \textbf{0.782} & \textbf{0.822} & \underline{0.786} & \underline{0.797}\\
        Llava (0-shot)   & 0.565 & 0.574 & 0.554 & 0.572\\
        Pixtral (0-shot) & 0.673 & 0.675 & 0.588 & 0.588\\
        \bottomrule
    \end{tabular}
    }
    \caption{(Macro-averaged) performance for the CheckThat! 2023 Task 1A benchmark. Best scores per sub-dataset are shown in \textbf{bold}, with the second-best \underline{underlined}.}
    \label{tab:checkthat20231a}
\end{table}

\begin{table*}
    \centering
    \begin{tabular}{@{}lccccccccc@{}}
        \toprule
        & 5Pils & \small Multiclaim & \small Flickr30K & \small SentiCap & \small Fakeddit& \multicolumn{4}{c}{Overall}\\
        \textbf{Model} & \textbf{Acc.} & \textbf{Acc.} & \textbf{Acc.} & \textbf{Acc.} & \textbf{Acc.} & \textbf{P.} & \textbf{R.} & \textbf{F1} & \textbf{Acc.}\\

        \midrule
        TinyBERT         & \underline{0.898} & \underline{0.878} & 0.794 & 0.721 & 0.480 & \underline{0.779} & \underline{0.796} & \underline{0.772} & \underline{0.775}\\
        BERT-base        & 0.682 & 0.611 & 0.573 & 0.816 & 0.803 & 0.669 & 0.675 & 0.671 & 0.685\\
        BERT-large       & 0.645 & 0.607 & 0.490 & \underline{0.854} & \underline{0.825} & 0.655 & 0.661 & 0.657 & 0.671\\
        ResNet           & 0.472 & 0.360 & 0.715 & 0.684 & 0.632 & 0.545 & 0.543 & 0.543 & 0.578\\
        ViT-base         & 0.321 & 0.353 & 0.710 & 0.689 & 0.766 & 0.529 & 0.526 & 0.525 & 0.571\\
        ViT-large        & 0.343 & 0.342 & 0.692 & 0.661 & 0.784 & 0.520 & 0.519 & 0.517 & 0.561\\
        BLIP             & 0.769 & 0.540 & \underline{0.934} & 0.600 & 0.416 & 0.656 & 0.661 & 0.657 & 0.671\\
        BLIP2            & 0.880 & 0.795 & 0.716 & 0.689 & 0.535 & 0.734 & 0.749 & 0.727 & 0.730\\
        Llava (0-shot)   & 0.225 & 0.266 & 0.053 & 0.449 & \textbf{0.996} & 0.328 & 0.318 & 0.320 & 0.334\\
        Pixtral (0-shot) & \textbf{0.954} & \textbf{0.919} & \textbf{0.937} & \textbf{0.962} & 0.716 & \textbf{0.909} & \textbf{0.921} & \textbf{0.914} & \textbf{0.918}\\
        \bottomrule
    \end{tabular}
    \caption{Test set of \hiot, performance per subset. The last four columns show the overall macro-averaged scores. Best scores per sub-dataset are shown in \textbf{bold}, with the second-best \underline{underlined}.}
    \label{tab:cross-dataset-perf}
\end{table*}

\subsection{Cross-modality Performance}
\label{sec:cross-modal}
On the CheckThat dataset, the strongest models are multimodal BLIP and BLIP2, which form the upper bound (see Table~\ref{tab:checkthat20231a}). Interestingly, the accuracies of text-only encoders are close to those of BLIP and BLIP2 (up to $94$\% relative to their accuracy), suggesting that little visual information is required for accurate checkworthiness detection, in line with results found in \citet{frick2023fraunhofer}. Image-only encoders also achieve only $14$\% lower accuracy than the upper bound. The narrow gap between single and multimodal models shows the dataset's limited suitability for assessing multimodal capabilities. ICL-based methods perform surprisingly poorly, with considerable false positive rates of $30$\% and $39$\% for Llava and Pixtral.

For the part of \hiot containing real data, Pixtral forms the upper performance bound (see Table~\ref{tab:cross-dataset-perf}). The gap between text-only models and the upper bound is larger than in CheckThat ($-14$\% vs.\ $-7$\%). Surprisingly, TinyBERT outperforms larger text-only models and is second to only Pixtral. This suggests that a small model with a well-tuned classification threshold can be effective, which poses an interesting venue for smaller organizations with limited compute capacity. Image-only encoders perform worse ($-35$\% vs.\ upper bound) than other models. Among multimodal models, BLIP2 excels, followed by Llava, aligning performance with parameter count (i.e., the bigger the model, the better its performance).

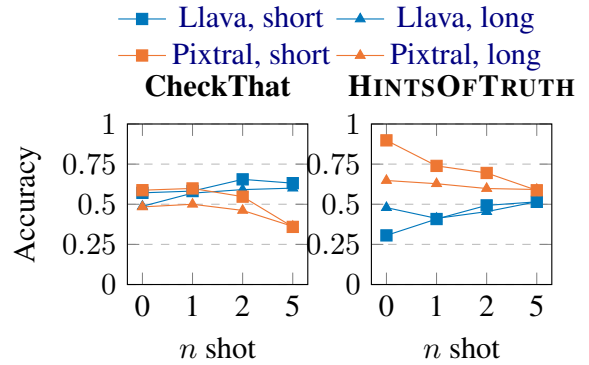
\begin{figure}[t]
    \centering
    \begin{tikzpicture}
    \begin{axis}[
        title={\textbf{CheckThat}},
        name={ct},
        xlabel={$n$ shot},
        ylabel={Accuracy},
        width=3.85cm,
        height=3.6cm,
        ymin=0, ymax=1,
        ytick={0,0.25,0.5,0.75,1},
        legend to name={legend-icl-perf},
        legend style={
            draw=none,
            fill=none,
        },
        legend columns=2, 
        xtick=data,
        ymajorgrids=true,
        grid style=dashed,
        symbolic x coords={0, 1, 2, 5}, 
    ]

    \addplot[
        color=ibm1,
        mark=square*,
    ]
    coordinates {
(0, 0.5711678832116789)
(1, 0.5821167883211679)
(2, 0.6551094890510949)
(5, 0.6313868613138686)
    };
    \addlegendentry{Llava, short}

    \addplot[
        color=ibm1,
        mark=triangle*,
    ]
    coordinates {
(0, 0.48722627737226276)
(1, 0.5693430656934306)
(2, 0.5912408759124088)
(5, 0.6003649635036497)
    };
    \addlegendentry{Llava, long}

    \addplot[
        color=ibm3,
        mark=square*,
    ]
    coordinates {
(0, 0.5875912408759124)
(1, 0.5985401459854015)
(2, 0.5474452554744526)
(5, 0.359)
    };
    \addlegendentry{Pixtral, short}

    \addplot[
        color=ibm3,
        mark=triangle*,
    ]
    coordinates {
(0, 0.4835766423357664)
(1, 0.5)
(2, 0.46167883211678834)
(5, 0.362)
    };
    \addlegendentry{Pixtral, long}
    \end{axis}

    \begin{axis}[
        at={($(ct.east)+(.8cm, 0)$)},
        name={ob},
        anchor=west,
        title={\textsc{\textbf{HintsOfTruth}}},
        xlabel={$n$ shot},
        width=3.85cm,
        height=3.6cm,
        ytick={0,0.25,0.5,0.75,1},
        ymin=0, ymax=1,
        legend pos=north west,
        xtick=data,
        ymajorgrids=true,
        grid style=dashed,
        symbolic x coords={0, 1, 2, 5}, 
    ]

    \addplot[
        color=ibm1,
        mark=square*,
    ]
    coordinates {
(0,0.306)
(1,0.408)
(2,0.493)
(5,0.515)
    };
    \addplot[
        color=ibm1,
        mark=triangle*,
    ]
    coordinates {
(0,0.479)
(1,0.412)
(2,0.454)
(5,0.516)
    };
    \addplot[
        color=ibm3,
        mark=square*,
    ]
    coordinates {
(0,0.898)
(1,0.739)
(2,0.695)
(5,0.587)
    };
    \addplot[
        color=ibm3,
        mark=triangle*,
    ]
    coordinates {
(0,0.648)
(1,0.628)
(2,0.598)
(5,0.592)
    };
    \end{axis}
    \node at ($(ct.east)!0.5!(ct.east)+(.31cm, 2.1)$) {\ref{legend-icl-perf}};
\end{tikzpicture}
    \caption{Few-shot performance with ICL on CheckThat (left) and \hiot (right).}
    \label{fig:icl-results-fewshot}
\end{figure}

We investigate the impact of $n$-shot learning on LLava and Pixtral by varying the prompting setup for these ICL models in Figure~\ref{fig:icl-results-fewshot}. Performance on \hiot reveals that Pixtral and Llava have contrary behavior with an increase in context:
\begin{enumerate*}[label=(\arabic*)]
    \item Adding more examples with few-shot learning aids Llava but hurts the Pixtral model, and
    \item Llava can benefit from a long prompt in zero-shot cases, but Pixtral generally benefits from short prompts.
\end{enumerate*} This is a surprising finding as additional examples should inform a model better. Like before, we observe an oversensitivity to predicting a \emph{checkworthy} label. The wide context of Pixtral may have it confuse which image/claim pair is currently under scrutiny.
While CheckThat shows this behavior partially, \hiot provides a clearer pattern, attesting to the usefulness of our dataset.

\subsection{Domain Generalization}
Evaluating performance on each of the \hiot subsets shows (see Table~\ref{tab:checkthat20231a}) that while fine-tuned (FT) models are trained on only the three domains using CheckThat, they consistently generalize to the subsets of \hiot, which suggests an effective knowledge transfer. However, performance varies based on experiment characteristics such as modalities used, pretraining setup, and model size.

\begin{figure*}[ht]
    \centering
    \begin{tikzpicture}
    \begin{groupplot}[
        group style={
            group name=my plots,
            group size=3 by 1,
            horizontal sep=1.4cm
        },
        ybar=2pt,
        /pgf/bar width=4.5pt,
        enlargelimits=0.15,
        enlarge y limits=0,
        scaled y ticks=false, 
        yticklabel style={/pgf/number format/.cd, fixed, precision=2},
        ylabel={PPR ($\uparrow$)},
        ymax=1.05,
        x tick label style={rotate=-45, anchor=north west, inner sep=0mm},
        symbolic x coords={TinyBERT, ResNet, BLIP2, Llava, Pixtral},
        xtick={TinyBERT, ResNet, BLIP2, Llava, Pixtral},
        xticklabels={TinyBERT, ResNet, BLIP2, Llava, Pixtral},
        legend columns=3,
        height=4cm,
        ymajorgrids=true,
        grid style=dashed,
        ymin=0,
        enlarge x limits=0.14,
        nodes near coords align={vertical},
        legend image code/.code={%
            \draw[#1] (0cm,-0.1cm) rectangle (0.6cm,0.1cm);
        }
    ]

    \nextgroupplot[
        title=\textbf{\footnotesize \makecell[c]{5Pils\\Checkworthy}},
        width=5.7cm
    ]
        \addplot[fill=ibm1!50, draw=ibm1] coordinates {
            (TinyBERT,0.916)
            (ResNet,0.423)
            (BLIP2,0.67065868263473)
            (Llava,0.17564870259481039)
            (Pixtral,0.7445109780439122)
        };
        \label{plot:series1}

        \addplot[fill=ibm4!50, draw=ibm4] coordinates { 
            (TinyBERT,0.916)
            (ResNet,0.560)
            (BLIP2,0.724) 
            (Llava,0.28)
            (Pixtral,0.66)
        };
        \label{plot:series4}
        \addplot[fill=ibm4!50, draw=ibm4,postaction={pattern=north west lines, pattern color=ibm4!70!black}] coordinates {
            (TinyBERT,0.916)
            (ResNet,0.482)
            (BLIP2,0.713)
            (Llava,0.319672131147541)
            (Pixtral,0.7438524590163934)
        };
        \label{plot:series5}

        \coordinate (legend) at (rel axis cs:1,1.38); 
        
    \nextgroupplot[
        title=\textbf{\footnotesize \makecell[c]{5Pils\\Non-checkworthy}},
        ylabel={PPR ($\downarrow$)},
        width=5.1cm,
    ]
        \addplot[fill=ibm3!50, draw=ibm3
        ] coordinates { 
            (TinyBERT,0.30538922155688625)
            (ResNet,0.423)
            (BLIP2,0.27944111776447106)
            (Llava,0.802) 
            (Pixtral,0.312)
        };

        \addplot[fill=ibm3!50, draw=ibm3,
            postaction={pattern=north east lines, pattern color=ibm3!70!black}
        ] coordinates {
            (TinyBERT,0.5808383233532934)
            (ResNet,0.423) 
            (BLIP2,0.3193612774451098)
            (Llava,0.4879032258064516)
            (Pixtral,0.3004032258064516)
        };

    \nextgroupplot[
        title=\textbf{\footnotesize \makecell[c]{Flickr30K\\Non-checkworthy}},
        width=8.7cm,
        ylabel={PPR ($\downarrow$)},
    ]
        
        \addplot[fill=ibm1!50, draw=ibm1] coordinates {(TinyBERT,0.5452685421994885) (ResNet,0.2751918158567775) (BLIP2,0.03682864450127877) (Llava,0.9432225063938618) (Pixtral,0.031713554987212275)};
        
        \addplot[fill=ibm3!50, draw=ibm3] coordinates {(TinyBERT,0.3943734015345269) (ResNet,0.2751918158567775) (BLIP2,0.043989769820971865) (Llava,0.9385560675883257) (Pixtral,0.01945724526369688)};
        \label{plot:series2}
        
        \addplot[fill=ibm3!50, draw=ibm3,             postaction={pattern=north east lines, pattern color=ibm3!70!black}] coordinates {(TinyBERT,0.5457800511508951) (ResNet,0.2751918158567775) (BLIP2,0.08951406649616368) (Llava,0.7892283790781979) (Pixtral,0.041429311237700675)};
        \label{plot:series3}

        \addplot[fill=ibm4!50, draw=ibm4] coordinates {(TinyBERT,0.5452685421994885) (ResNet,0.451150895140665) (BLIP2,0.0782608695652174) (Llava,0.9601023017902813) (Pixtral,0.043478260869565216)};
        
        \addplot[fill=ibm4!50, draw=ibm4,postaction={pattern=north west lines, pattern color=ibm4!70!black}] coordinates {(TinyBERT,0.545) (ResNet,0.415) (BLIP2,0.0859) (Llava,0.9580562659846548) (Pixtral,0.05421994884910486)};
    \end{groupplot}

    \node at (legend) [anchor=base, xshift=4.7cm, yshift=0.3cm] {
        \begin{tabular}{l}
            \ref{plot:series1} Real 
            \ref{plot:series2} Text (LL)
            \ref{plot:series3} Text (B)
            \ref{plot:series4} Img (F)
            \ref{plot:series5} Img (SD)
        \end{tabular}
    };

\end{tikzpicture}
    \caption{Positive prediction rate (PPR) for each model on the two augmented subsets in \hiot compared to real-world data, with 5Pils plot split into checkworthy and non-checkworthy samples. $\uparrow$ and $\downarrow$ denote higher and lower as better, respectively. \textbf{Text (LL)}: Llava-generated caption, \textbf{Text (B)}: BLIP-generated caption, \textbf{Img (F)}: Flux-generated image, \textbf{Img (SD)}: StableDiffusion-generated image.}
    \label{fig:synthetic-augmentation}
\end{figure*}
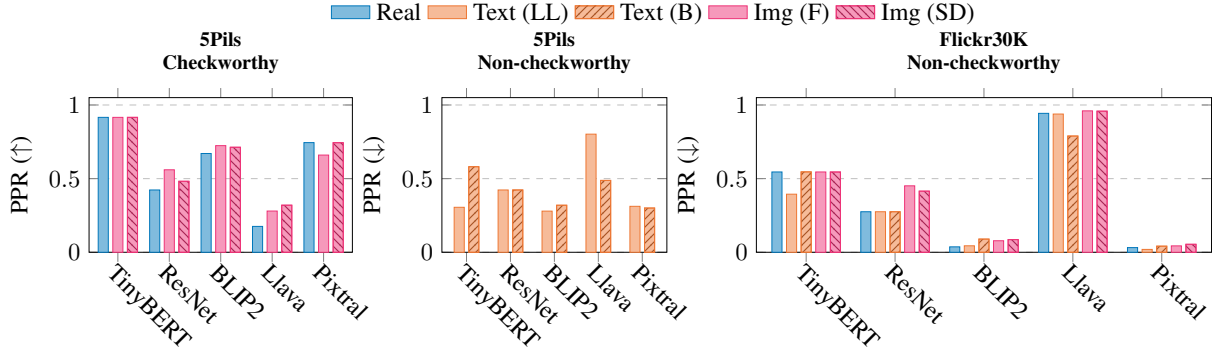

Among FT models, TinyBERT is robust across most datasets but struggles on Fakeddit, likely due to the linguistic differences between CheckThat and Fakeddit; Text data in the latter stems from user-submitted post titles, which are less grammatically correct.\footnote{Example: \emph{``took this photo of my dog rolling in some grass''} for Fakeddit vs.\ \emph{``a photograph shows rays of lights in the shape of a cross during the august 2017 eclipse.''} for Multiclaim.} Larger BERT models perform well on Fakeddit but worse on SentiCap and Multiclaim. Since TinyBERT is distilled from these models, constraining model size may enhance generalization but influence error modes.

ICL performance also varies: Llava achieves the highest accuracy on Fakeddit, while Pixtral excels on all other subsets. Llava’s training on noisy user-generated ShareGPT4V data \citep{chen2024sharegpt4v} may explain its behavior, while Pixtral may favor syntactically correct texts. This difference between the two models highlights noisy data as a unique generalization challenge. Finally, BLIP excels on Flickr30K, despite not being finetuned on it, raising data leakage concerns \citep{balloccu2024leak}.

\subsection{Performance on Synthetic Data}
\begin{figure*}[t]
\centering
\begin{tikzpicture}
    \matrix[matrix of nodes, nodes={inner sep=0,anchor=center}, column sep=1mm, row sep=1mm] (m) {
        |[label={[text width=0.37\columnwidth, align=center]above:\textbf{Original}}]| \includegraphics[width=0.37\columnwidth, trim=34 4 55 4 px, clip]{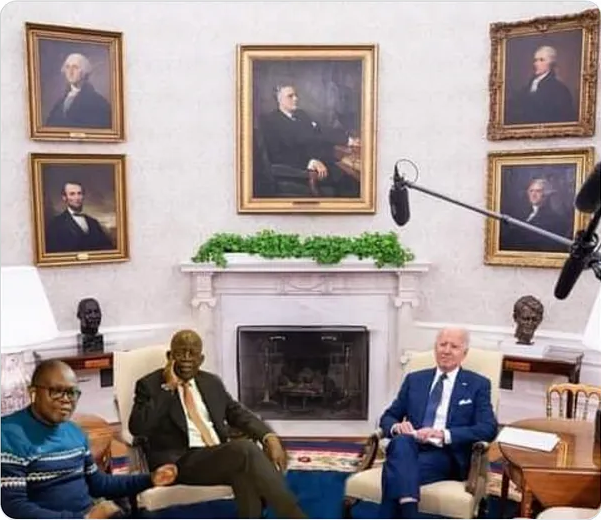} &
        |[label={[text width=0.37\columnwidth, align=center]above:\textbf{Flux-generated}}]|\includegraphics[width=0.37\columnwidth]{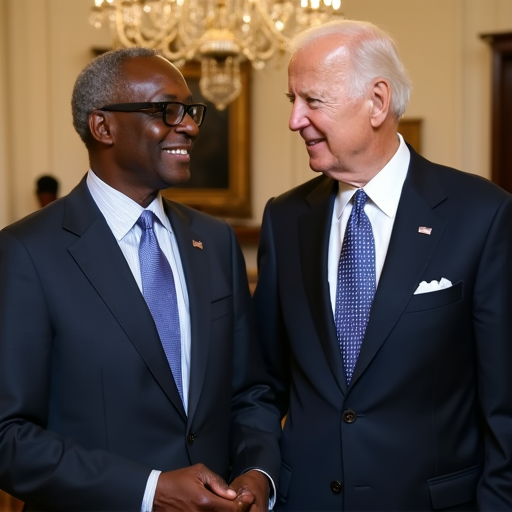} &
        |[label={[text width=0.37\columnwidth, align=center]above:\textbf{SD3.5-generated }}]|\includegraphics[width=0.37\columnwidth]{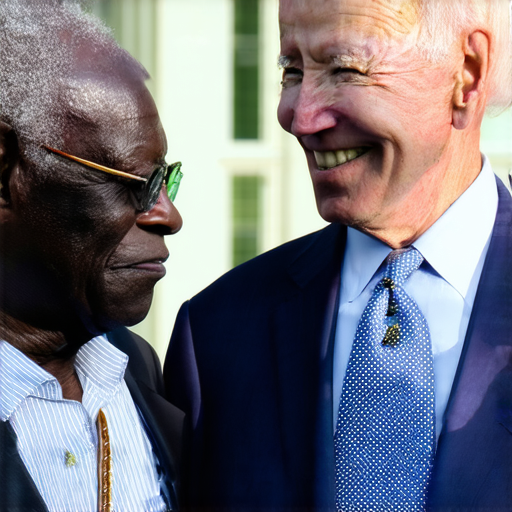} \\
    };
    \node[below=4mm of current bounding box.south, text width=1\columnwidth, align=center, draw=black] (original-caption) {
        \small{\emph{``The photo shows Nigeria’s presidential candidate Bola Tinubu with US President Joe Biden at the White House.''}}
    };
    \node[above=0mm of original-caption, yshift=-1mm, text width=0.66\columnwidth, align=center] (original-title) {
        \textbf{Original caption}
    };
    \node[right=2mm of m-1-3, text width=0.72\columnwidth, align=center, draw=black] (llama-caption) {
        \small{\emph{``Three men are sitting in a room, with one man wearing a suit and tie, another wearing a suit and a blue shirt, and the third man wearing a suit and <truncated for brevity>''}}
    };
    \node[above=0mm of llama-caption, yshift=-1mm, text width=0.77\columnwidth, align=center] (llama-title) {
        \textbf{Llava-generated caption}
    };
    \node[below=7mm of llama-caption, text width=0.72\columnwidth, align=center, draw=black] (blip-caption) {
        \small{\emph{``Image of a man in a suit and tie sitting in a chair''}}
    };
    \node[above=0mm of blip-caption, yshift=-1mm, text width=0.77\columnwidth, align=center] (blip-title) {
        \textbf{BLIP-generated caption}
    };

    \matrix[below=0mm of original-caption, matrix of nodes, nodes={inner sep=0,anchor=center}, column sep=1mm, row sep=1mm] (m-flickr) {
        |[label={[text width=0.37\columnwidth, align=center]above:\textbf{Original}}]| \includegraphics[width=0.37\columnwidth, trim=125 0 44 0 px, clip]{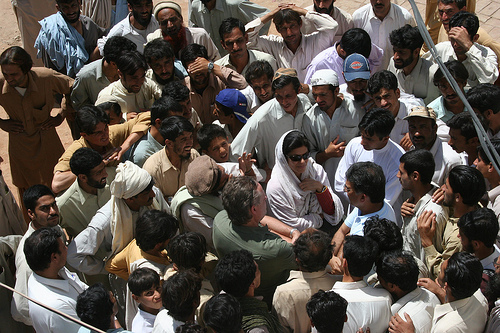} &
        |[label={[text width=0.37\columnwidth, align=center]above:\textbf{Flux-generated}}]|\includegraphics[width=0.37\columnwidth]{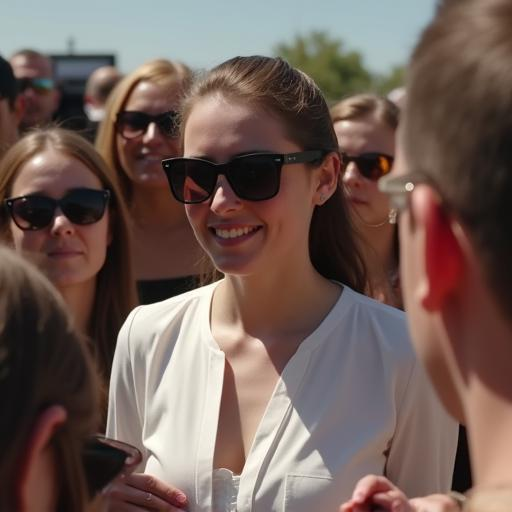} &
        |[label={[text width=0.37\columnwidth, align=center]above:\textbf{SD3.5-generated}}]|\includegraphics[width=0.37\columnwidth]{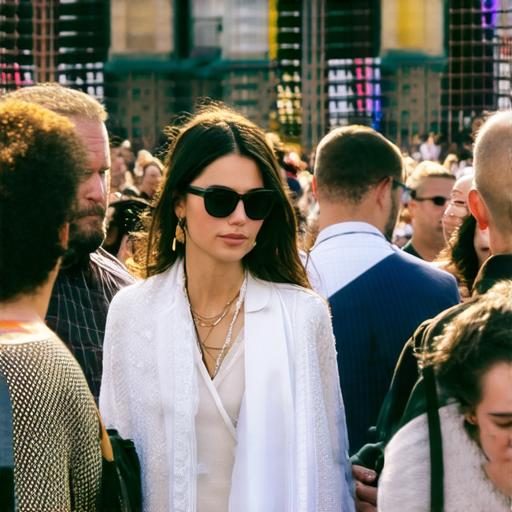}\\
    };
    \node[below=4mm of m-flickr.south, draw=black, text width=1\columnwidth, align=center] (original-caption-flickr) {
        \small{\emph{``A crowd of people surrounding a woman in white wearing sunglasses.''}}
    };
    \node[above=0mm of original-caption-flickr, yshift=-1mm, text width=0.77\columnwidth, align=center] (original-title-flickr) {
        \textbf{Original caption}
    };
    \node[right=2mm of m-flickr-1-3, text width=0.72\columnwidth, align=center, draw=black] (llama-caption-flickr) {
        \small{\emph{``A diverse group of people is gathered together in a crowd, with a woman in a headscarf being held by a man in a green shirt.''}}
    };
    \node[above=0mm of llama-caption-flickr, yshift=-1mm, text width=0.77\columnwidth, align=center] (llama-title-flickr) {
        \textbf{Llava-generated caption}
    };
    \node[below=7mm of llama-caption-flickr, text width=0.72\columnwidth, align=center, draw=black] (blip-caption-flickr) {
        \small{\emph{``There are many people standing around a man in a crowd''}}
    };
    \node[above=0mm of blip-caption-flickr, yshift=-1mm, text width=0.77\columnwidth, align=center] (blip-title-flickr) {
        \textbf{BLIP-generated caption}
    };

\end{tikzpicture}
\caption{Examples of synthetically generated images and captions. The upper row shows a checkworthy example from 5Pils. The bottom row shows a non-checkworthy example from Flickr30K.}
\label{tab:synthetic-examples}
\end{figure*}
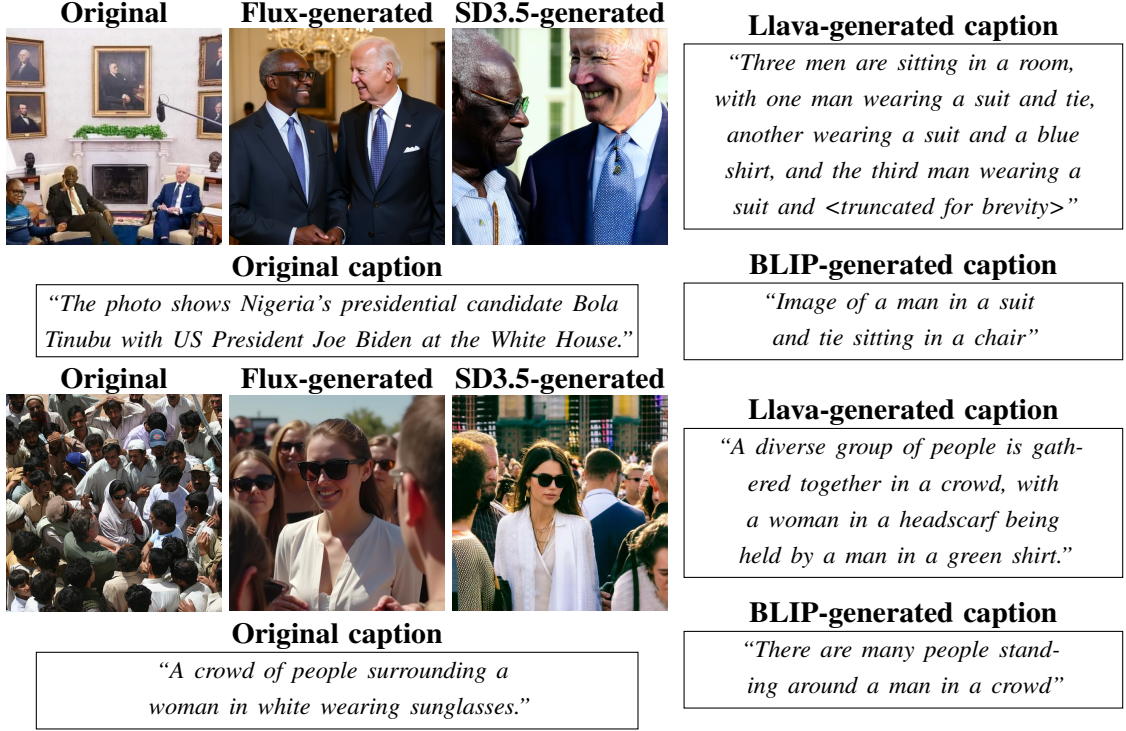

We investigate prediction behavior on the synthetic part of \hiot, using images generated by Flux \citep{flux2024} and Stable Diffusion 3.5 \citep{stablediff}, and textual claims by Llava \citep{li2024llavanext} and BLIP \citep{li2022blip}. To the human eye, synthetic samples appear distinct from real-world samples (see Figure~\ref{tab:synthetic-examples} for some examples). Our goal is to determine whether models can reliably detect synthetic data and differentiate between various generative methods. To achieve this, we cross-check with the same models used for classification to assess whether they can identify their own synthetic generations. We evaluate a subset of models, including the smallest (TinyBERT, ResNet) and largest (BLIP2, Llava, Pixtral), to analyze compute/accuracy tradeoffs. Figure~\ref{fig:synthetic-augmentation} provides an overview of the positive checkworthiness prediction rate (PPR) per model to reveal how frequently each model classifies an image/claim pair as checkworthy, shedding light on both model biases toward synthetic modalities and their potential failure modes.

\paragraph{Results}
TinyBERT is accurate on real 5Pils data but struggles with synthetic text. For example, it misclassifies over half of BLIP-generated texts as checkworthy. It also has a high false positive rate ($\sim0.55$) on augmentations of Flickr30K, which could lead to an unnecessarily high workload for fact-checkers. ResNet, on the other hand, often misses checkworthy samples for 5Pils (high false negative rate). It's higher PPR for synthetic images--by $32$\% for Flux and by $14$\% for SD on 5Pils and $64$\% Flux and $51$\% for SD on Flickr30K shows that ResNet may be overly sensitive to detecting synthetic images as checkworthy, even if the synthetic images are not relevant (see Section~\ref{sec:method-generating}).

Llava generates many false negatives on 5Pils while obtaining a high false positive rate on Flickr30K, suggesting that the model may misunderstand the task instructions. The high PPR on Llava-generated texts for 5Pils reveals an oversensitivity to synthetic texts generated by itself.
BLIP2 behaved more in line with expectations, with a lower PPR for synthetic texts while sustaining a high PPR for synthetic images in 5Pils. On Flickr30K, it maintained a minimal PPR across all synthetic data, likely benefiting from pretraining on synthetic captions. The origin of the synthetic text (BLIP vs.\ Llava) had little impact on its performance. Pixtral mirrors BLIP2's results, except that pairs from 5Pils with images generated by Flux were $10$\% less often identified as checkworthy by Pixtral, suggesting that as newer, higher-quality image generators emerge, Pixtral’s accuracy might decline.

\subsection{Compute Budget}
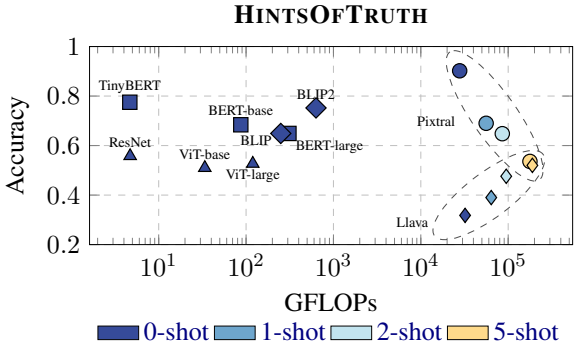
\begin{figure}[ht]
    \centering
    \begin{tikzpicture}

\begin{axis}[
    title={\textsc{\textbf{HintsOfTruth}}},
    name={budget-plot},
    xlabel={GFLOPs},
    width=8.7cm,
    height=4.5cm,
    ymajorgrids=true,
    xmajorgrids=true,
    grid style=dashed,
    legend to name=compute-legend,
    legend columns=4,
    legend style={
        draw=none,
    },
    ymin=0,
    xmode=log,
    ymax=1,
    ymin=0.2,
    ylabel={Accuracy},
    enlarge x limits=0.1,
    scatter/classes={
        0={mark=square*, rotate=45,   draw=black, fill=grad1},
        1={mark=square*,    draw=black, fill=grad1},
        2={mark=triangle*,  draw=black, fill=grad1},
        3={mark=diamond*,   draw=black, fill=grad1},
        4={mark=diamond*,   draw=black, fill=grad2},
        5={mark=diamond*,   draw=black, fill=grad3},
        6={mark=diamond*,   draw=black, fill=grad4},
        7={mark=*,  draw=black, fill=grad1},
        8={mark=*,  draw=black, fill=grad2},
        9={mark=*,  draw=black, fill=grad3},
        10={mark=*, draw=black, fill=grad4}
    }
]
\addlegendimage{area legend, 
               fill=grad1,
               draw=black,
               minimum width=3mm,
               minimum height=2mm}
\addlegendentry{0-shot}
\addlegendimage{area legend, 
               fill=grad2,
               draw=black,
               minimum width=3mm,
               minimum height=2mm}
\addlegendentry{1-shot}
\addlegendimage{area legend, 
               fill=grad3,
               draw=black,
               minimum width=3mm,
               minimum height=2mm}
\addlegendentry{2-shot}
\addlegendimage{area legend, 
               fill=grad4,
               draw=black,
               minimum width=3mm,
               minimum height=2mm}
\addlegendentry{5-shot}

\addplot[
    scatter, only marks,
    scatter src=explicit symbolic,
    mark size=3pt,
    mark options={line width=0.5pt},
] table[
    meta=model_type,                     
    x=x,
    y=y,
    header=true                          
] {
    x    y    model_type    model_name
    4.67  0.775  1  TinyBERT
    87.02 0.684  1  BERT-base
    309.37 0.649 1  BERT-large
    4.71   0.558 2  ResNet
    33.71  0.510 2  ViT-base
    119.34 0.527 2  ViT-large
    249.58 0.649 0  BLIP
    631.59 0.752 0  BLIP2
    27843  0.902 7  Pixtral
    55943  0.690 8  Pixtral
    85630  0.648 9  Pixtral
    177391 0.536 10  Pixtral
    32117  0.318 3  Llava
    63963  0.390 4  Llava
    94930  0.476 5  Llava
    189197 0.520 6  Llava

};

\addplot+[
    scatter, only marks,
    scatter src=explicit symbolic,
    mark size=0pt,
    nodes near coords,                   
    nodes near coords align={above},     
    point meta=explicit symbolic,        
    every node near coord/.style={
        anchor=south,
        font=\tiny,
        black
    },
] table[
    meta=model_name,                     
    col sep=space,                       
    x=x,
    y=y,
    header=true                          
] {
    x    y       model_name
    4.67   0.775 TinyBERT
    87.02  0.684 BERT-base
    309.37 0.649 {}
    4.71   0.558 ResNet
    33.71  0.510 ViT-base
    119.34 0.527 {}
    249.58 0.649 {}
    631.59 0.752 BLIP2
    
};

\node[] at (axis cs: 900, 0.595) {\tiny BERT-large};
\node[] at (axis cs: 130, 0.62) {\tiny BLIP};
\node[] at (axis cs: 119.34, 0.477) {\tiny ViT-large};
\node[ellipse,draw=black!70, dashed,minimum width=2cm, minimum height=.66cm,rotate=37] at (axis cs: 60000, 0.40) {};
\node[ellipse,draw=black!70, dashed,minimum width=2.3cm, minimum height=.66cm,rotate=125] at (axis cs: 70000, 0.72) {};
\node[] at (axis cs: 8000, 0.29) {\tiny Llava};
\node[] at (axis cs: 15000, 0.70) {\tiny Pixtral};

\end{axis}
\node[anchor=base] at (3.55cm,-1.6cm) {\ref{compute-legend}};

\end{tikzpicture}
    \caption{Compute budget versus task performance.}
    \label{fig:compute-budget}
\end{figure}

Unsurprisingly, models with more parameters generally perform better at checkworthiness detection. However, running large models like Pixtral demands substantial compute resources. Since checkworthiness detection serves as a prefiltering task, such resources may not be available to media organizations or outpaced by new content. To explore the trade-off between model size and performance, we visualize the compute budget in FLOPs \citep{hassid2024larger} compared to final accuracy in Figure~\ref{fig:compute-budget}. FLOPs usage and wall time are estimated using the \texttt{calflops} library \citep{calflops}, averaging over 100 random samples from \hiot, measured on a node with a single H100 GPU.


\paragraph{Results}
The best-performing model, Pixtral, requires at least two orders of magnitude more compute than FT models, even with zero-shot ICL. BLIP2 offers a balanced trade-off, ranking third in accuracy at a reasonable compute cost. However, in wall time, it closely matches ICL models---on average, BLIP2 runs as long as 1-shot Pixtral, while Pixtral 0-shot is up to 36\% faster per sample (see App~\ref{app:wall-time} for details). TinyBERT emerges as the most balanced, delivering competitive accuracy at significantly lower cost and runtime (four orders of magnitude in FLOPS, two in wall time). This suggests that tuning a small model can achieve strong performance, raising questions about the role of visual information in checkworthiness detection.

\section{Conclusions}
\hiot provides key insights into the challenges and opportunities in multimodal checkworthiness detection and the questionable role that visual content plays in misinformation.
Our findings indicate that while multimodal models outperform image-only approaches, their advantage over text-only models is not attested. Well-tuned text-based models achieve nearly the same accuracy (up to 86\%), raising uncertainty about \emph{the extent to which visual content contributes} to the checkworthiness of real-world image/claim pairs.
Unlike many other areas of NLP, our experiments reveal that the \emph{syntactic} and \emph{grammatical structure} of the claims, rather than their domain, impacts generalization. Larger models, like Pixtral, demonstrate high adaptability but may unexpectedly fail to transfer.
When confronted with synthetic data, lightweight models become \emph{oversensitive}, often misclassifying images as checkworthy. This increases fact-checkers' workload by requiring manual filtering of false positives. Fine-tuning models on synthetic samples could help, but risks turning into an adversarial race with evolving image generators \citep{corvi2023detection}.
Our analysis of the computational trade-offs reveals that large models come at compute costs of \emph{four orders of magnitude larger} than smaller models like TinyBERT. Though small models require careful tuning to be conservative, their lightweight nature makes them practically be better suited as checkworthiness detection methods.

Future work should shift checkworthiness detection to a ranking-based approach, helping fact-checkers prioritize claims. Explain why a claim needs verification can further help fact-checkers communicate decisions \citep{mccright2017combatting}, for instance, by collaboratively uncovering the arguments supporting a fact-check decision \citep{vandermeer2024hybrid}. Techniques like Learning to Defer \citep{madras2018predict, khurana2024crowd} and Active Learning \citep{vandermeer2024annotator} assist in efficient data collection.

\section*{Limitations}
Several limitations have an impact on the findings of our work. First, our method of checking for complex image use by retaining only claims that mention multimodal content, to answer \textbf{Q4} (Does the image contribute extra information to the claim?), can be strengthened. Specifically, for the checkworthy subsets (5Pils and Multiclaim), we assume that, since the claims and images were included in a fact-checking article, the image provides additional context as part of the fact-check. However, further (manual) verification would be necessary to test this assumption empirically.
Similarly, we assume that the image generator will generate appropriate context when considering the prompt for the augmented versions of these subsets. Again, this is a strong assumption, as generated images can be said to provide (potentially irrelevant) context. Nonetheless, we believe the augmentations to be checkworthy: models can learn from visual artifacts or types of generated contexts.

Second, our study is conducted entirely on English data, whereas misinformation has impacts across many different languages and cultures. However, some of the resources used in our work could be exploited to generate instances in other languages.

Third, we do not incorporate retrieval-augmented generation (RAG) systems in our experiments. While such systems could potentially enhance checkworthiness detection by retrieving relevant fact-checks \citep{singal2024evidence}, they are sensitive to temporal leakage when past fact-checks are accessible \citep{glockner2022missing}, skewing the results, and require even further resources than the models in this paper.

Finally, we do not conduct a human evaluation of checkworthiness predictions. While crowd annotators are often employed for such tasks, their ability to accurately judge checkworthiness remains uncertain. Fact-checking services often employ expert journalists who draw on their intuition and experience to decide what to fact-check and may take up to a couple of days to write fact-checking articles. Whether lay crowd annotators can reliably annotate checkworthiness in an online annotation study is therefore unclear. Parallel crowd and expert evaluation studies, such as expert assessments or real-world fact-checking use cases, could provide deeper insights into annotator behavior.

\section*{Ethical Considerations}
The development of multimodal fact-checking datasets, like HintsOfTruth, involves several critical ethical considerations to ensure societal benefit.

\paragraph{Bias Mitigation}
Data is sourced from diverse domains, including social media (e.g., Multiclaim) and datasets that focus on underrepresented cultures (e.g., 5Pils). However, since we primarily reuse existing datasets, our corpus remains limited in size and inclusivity. As a result, geographic and cultural biases may persist.

\paragraph{Anonymization}
To protect user privacy, real-world data from social media and fact-checking articles is anonymized. Image/claim pairs are stripped of personally identifiable information (PII), and no additional contextual information is introduced. We adhere strictly to the licensing terms of the publicly available datasets we use.

\paragraph{Misinformation Risks}
As our system contributes to the fact-checking pipeline, it is designed to help combat misinformation. However, synthetic data generation tools have the potential for misuse. To mitigate this risk, our study explicitly avoids introducing adversarial prompts that could be exploited for harmful purposes.

\paragraph{Resource Accessibility}
We prioritize lightweight models, such as TinyBERT, to enhance scalability and ensure that organizations with limited computational resources can access misinformation detection tools. Additionally, all models used in our research, including the largest ICL models, are freely available on the HuggingFace Hub.

\section*{Acknowledgements}
This research was sponsored by Hasler Foundation’s FactCheck project. This work was performed using the compute resources from the Idiap Research Institute and the Academic Leiden Interdisciplinary Cluster Environment (ALICE) provided by Leiden University. We would also like to thank the ARR
reviewers for their helpful feedback.

\bibliography{compact.fixed}

\begin{thebibliography}{84}
\providecommand{\natexlab}[1]{#1}

\bibitem[{Abdelnabi et~al.(2022)Abdelnabi, Hasan, and
  Fritz}]{abdelnabi2022open}
Sahar Abdelnabi, Rakibul Hasan, and Mario Fritz. 2022.
\newblock Open-domain, content-based, multi-modal fact-checking of
  out-of-context images via online resources.
\newblock In \emph{Proceedings of the IEEE/CVF Conference on Computer Vision
  and Pattern Recognition (CVPR)}, pages 14940--14949.

\bibitem[{Akata et~al.(2020)Akata, Balliet, de~Rijke, Dignum, Dignum, Eiben,
  Fokkens, Grossi, Hindriks, Hoos, Hung, Jonker, Monz, Neerincx, Oliehoek,
  Prakken, Schlobach, van~der Gaag, van Harmelen, van Hoof, van Riemsdijk, van
  Wynsberghe, Verbrugge, Verheij, Vossen, and Welling}]{akata2020research}
Zeynep Akata, Dan Balliet, Maarten de~Rijke, Frank Dignum, Virginia Dignum,
  Guszti Eiben, Antske Fokkens, Davide Grossi, Koen~V. Hindriks, Holger~H.
  Hoos, Hayley Hung, Catholijn~M. Jonker, Christof Monz, Mark~A. Neerincx,
  Frans~A. Oliehoek, Henry Prakken, Stefan Schlobach, Linda~C. van~der Gaag,
  Frank van Harmelen, Herke van Hoof, Birna van Riemsdijk, Aimee van
  Wynsberghe, Rineke Verbrugge, Bart Verheij, Piek Vossen, and Max Welling.
  2020.
\newblock \href {https://doi.org/10.1109/MC.2020.2996587} {{A} {R}esearch
  {A}genda for {H}ybrid {I}ntelligence: {A}ugmenting {H}uman {I}ntellect {W}ith
  {C}ollaborative, {A}daptive, {R}esponsible, and {E}xplainable {A}rtificial
  {I}ntelligence}.
\newblock \emph{Computer}, 53(8):18--28.

\bibitem[{Akhtar et~al.(2023)Akhtar, Schlichtkrull, Guo, Cocarascu, Simperl,
  and Vlachos}]{akhtar-etal-2023-multimodal}
Mubashara Akhtar, Michael Schlichtkrull, Zhijiang Guo, Oana Cocarascu, Elena
  Simperl, and Andreas Vlachos. 2023.
\newblock \href {https://doi.org/10.18653/v1/2023.findings-emnlp.361}
  {Multimodal automated fact-checking: A survey}.
\newblock In \emph{Findings of the Association for Computational Linguistics:
  EMNLP 2023}, pages 5430--5448, Singapore. Association for Computational
  Linguistics.

\bibitem[{Alam et~al.(2023)Alam, Barr{\'o}n-Cede{\~n}o, Cheema, Shahi, Hakimov,
  Hasanain, Li, M{\'\i}guez, Mubarak, Zaghouani et~al.}]{alam2023overview}
Firoj Alam, Alberto Barr{\'o}n-Cede{\~n}o, Gullal~S Cheema, Gautam~Kishore
  Shahi, Sherzod Hakimov, Maram Hasanain, Chengkai Li, Rub{\'e}n M{\'\i}guez,
  Hamdy Mubarak, Wajdi Zaghouani, et~al. 2023.
\newblock Overview of the clef-2023 checkthat! lab task 1 on check-worthiness
  of multimodal and multigenre content.

\bibitem[{Augenstein et~al.(2024)Augenstein, Baldwin, Cha, Chakraborty,
  Ciampaglia, Corney, DiResta, Ferrara, Hale, Halevy, Hovy, Ji, Menczer,
  M{\'{\i}}guez, Nakov, Scheufele, Sharma, and
  Zagni}]{augenstein2024factuality}
Isabelle Augenstein, Timothy Baldwin, Meeyoung Cha, Tanmoy Chakraborty,
  Giovanni~Luca Ciampaglia, David P.~A. Corney, Renee DiResta, Emilio Ferrara,
  Scott Hale, Alon~Y. Halevy, Eduard~H. Hovy, Heng Ji, Filippo Menczer,
  Rub{\'{e}}n M{\'{\i}}guez, Preslav Nakov, Dietram Scheufele, Shivam Sharma,
  and Giovanni Zagni. 2024.
\newblock \href {https://doi.org/10.1038/S42256-024-00881-Z} {{F}actuality
  challenges in the era of large language models and opportunities for
  fact-checking}.
\newblock \emph{Nat. Mac. Intell.}, 6(8):852--863.

\bibitem[{Balloccu et~al.(2024)Balloccu, Schmidtov{\'a}, Lango, and
  Dusek}]{balloccu2024leak}
Simone Balloccu, Patr{\'i}cia Schmidtov{\'a}, Mateusz Lango, and Ondrej Dusek.
  2024.
\newblock \href {https://aclanthology.org/2024.eacl-long.5/} {Leak, cheat,
  repeat: Data contamination and evaluation malpractices in closed-source
  {LLM}s}.
\newblock In \emph{Proceedings of the 18th Conference of the European Chapter
  of the Association for Computational Linguistics (Volume 1: Long Papers)},
  pages 67--93, St. Julian{'}s, Malta. Association for Computational
  Linguistics.

\bibitem[{Barr{\'o}n-Cedeno et~al.(2020)Barr{\'o}n-Cedeno, Elsayed, Nakov,
  Da~San~Martino, Hasanain, Suwaileh, Haouari, Babulkov, Hamdan, Nikolov
  et~al.}]{barron2020overview}
Alberto Barr{\'o}n-Cedeno, Tamer Elsayed, Preslav Nakov, Giovanni
  Da~San~Martino, Maram Hasanain, Reem Suwaileh, Fatima Haouari, Nikolay
  Babulkov, Bayan Hamdan, Alex Nikolov, et~al. 2020.
\newblock Overview of checkthat! 2020: Automatic identification and
  verification of claims in social media.
\newblock In \emph{Experimental IR Meets Multilinguality, Multimodality, and
  Interaction: 11th International Conference of the CLEF Association, CLEF
  2020, Thessaloniki, Greece, September 22--25, 2020, Proceedings 11}, pages
  215--236. Springer.

\bibitem[{{Black Forest Labs}(2024)}]{flux2024}
{Black Forest Labs}. 2024.
\newblock {FLUX.1: {A} {T}ext-to-Image {G}eneration {M}odel}.
\newblock https://blackforestlabs.ai.
\newblock Accessed January 16, 2025.

\bibitem[{Bonet{-}Jover et~al.(2024)Bonet{-}Jover, Sep{\'{u}}lveda{-}Torres,
  Saquete, Mart{\'{\i}}nez{-}Barco, and Nieto{-}P{\'{e}}rez}]{bonet2024run}
Alba Bonet{-}Jover, Robiert Sep{\'{u}}lveda{-}Torres, Estela Saquete, Patricio
  Mart{\'{\i}}nez{-}Barco, and Mario Nieto{-}P{\'{e}}rez. 2024.
\newblock \href {https://doi.org/10.1007/S10579-023-09678-9} {{RUN-AS:} a novel
  approach to annotate news reliability for disinformation detection}.
\newblock \emph{Lang. Resour. Evaluation}, 58(2):609--639.

\bibitem[{Byeon et~al.(2022)Byeon, Park, Kim, Lee, Baek, and
  Kim}]{kakaobrain2022coyo-700m}
Minwoo Byeon, Beomhee Park, Haecheon Kim, Sungjun Lee, Woonhyuk Baek, and
  Saehoon Kim. 2022.
\newblock {C}{O}{Y}{O}-700{M}: {I}mage-text {P}air {D}ataset.
\newblock \url{https://github.com/kakaobrain/coyo-dataset}.

\bibitem[{Cheema et~al.(2022)Cheema, Hakimov, Sittar, M{\"u}ller-Budack, Otto,
  and Ewerth}]{cheema2022mm}
Gullal~Singh Cheema, Sherzod Hakimov, Abdul Sittar, Eric M{\"u}ller-Budack,
  Christian Otto, and Ralph Ewerth. 2022.
\newblock \href {https://doi.org/10.18653/v1/2022.findings-naacl.72}
  {{MM}-claims: A dataset for multimodal claim detection in social media}.
\newblock In \emph{Findings of the Association for Computational Linguistics:
  NAACL 2022}, pages 962--979, Seattle, United States. Association for
  Computational Linguistics.

\bibitem[{Chen and Shu(2024)}]{chen2024combating}
Canyu Chen and Kai Shu. 2024.
\newblock Combating misinformation in the age of llms: Opportunities and
  challenges.
\newblock \emph{AI Magazine}, 45(3):354--368.

\bibitem[{Chen et~al.(2022)Chen, Sriram, Choi, and
  Durrett}]{chen2022generating}
Jifan Chen, Aniruddh Sriram, Eunsol Choi, and Greg Durrett. 2022.
\newblock \href {https://doi.org/10.18653/v1/2022.emnlp-main.229} {Generating
  literal and implied subquestions to fact-check complex claims}.
\newblock In \emph{Proceedings of the 2022 Conference on Empirical Methods in
  Natural Language Processing}, pages 3495--3516, Abu Dhabi, United Arab
  Emirates. Association for Computational Linguistics.

\bibitem[{Chen et~al.(2024)Chen, Li, Dong, Zhang, He, Wang, Zhao, and
  Lin}]{chen2024sharegpt4v}
Lin Chen, Jinsong Li, Xiaoyi Dong, Pan Zhang, Conghui He, Jiaqi Wang, Feng
  Zhao, and Dahua Lin. 2024.
\newblock \href {https://sharegpt4v.github.io//} {Sharegpt4v: Improving large
  multi-modal models with better captions}.
\newblock In \emph{European Conference on Computer Vision}, pages 370--387.
  Springer.

\bibitem[{Chen et~al.(2021)Chen, Xiao, and Mao}]{chen2021persuasion}
Sijing Chen, Lu~Xiao, and Jin Mao. 2021.
\newblock \href {https://doi.org/10.1016/J.IPM.2021.102665} {{P}ersuasion
  strategies of misinformation-containing posts in the social media}.
\newblock \emph{Inf. Process. Manag.}, 58(5):102665.

\bibitem[{Corvi et~al.(2023)Corvi, Cozzolino, Zingarini, Poggi, Nagano, and
  Verdoliva}]{corvi2023detection}
Riccardo Corvi, Davide Cozzolino, Giada Zingarini, Giovanni Poggi, Koki Nagano,
  and Luisa Verdoliva. 2023.
\newblock \href {https://ieeexplore.ieee.org/abstract/document/10095167} {On
  the detection of synthetic images generated by diffusion models}.
\newblock In \emph{ICASSP 2023-2023 IEEE International Conference on Acoustics,
  Speech and Signal Processing (ICASSP)}, pages 1--5. IEEE.

\bibitem[{Das et~al.(2023)Das, Liu, Kovatchev, and Lease}]{das2023state}
Anubrata Das, Houjiang Liu, Venelin Kovatchev, and Matthew Lease. 2023.
\newblock The state of human-centered nlp technology for fact-checking.
\newblock \emph{Information processing \& management}, 60(2):103219.

\bibitem[{Del~Vicario et~al.(2016)Del~Vicario, Bessi, Zollo, Petroni, Scala,
  Caldarelli, Stanley, and Quattrociocchi}]{del2016spreading}
Michela Del~Vicario, Alessandro Bessi, Fabiana Zollo, Fabio Petroni, Antonio
  Scala, Guido Caldarelli, H~Eugene Stanley, and Walter Quattrociocchi. 2016.
\newblock {T}he spreading of misinformation online.
\newblock \emph{Proceedings of the national academy of Sciences},
  113(3):554--559.

\bibitem[{Dong et~al.(2024)Dong, Li, Dai, Zheng, Ma, Li, Xia, Xu, Wu, Chang,
  Sun, and Sui}]{dong2022survey}
Qingxiu Dong, Lei Li, Damai Dai, Ce~Zheng, Jingyuan Ma, Rui Li, Heming Xia,
  Jingjing Xu, Zhiyong Wu, Baobao Chang, Xu~Sun, and Zhifang Sui. 2024.
\newblock \href {https://aclanthology.org/2024.emnlp-main.64} {{A} {S}urvey on
  {I}n-context {L}earning}.
\newblock In \emph{Proceedings of the 2024 Conference on Empirical Methods in
  Natural Language Processing, {EMNLP} 2024, Miami, FL, USA, November 12-16,
  2024}, pages 1107--1128. Association for Computational Linguistics.

\bibitem[{Dufour et~al.(2024)Dufour, Pathak, Samangouei, Hariri, Deshetti,
  Dudfield, Guess, Escayola, Tran, Babakar, and Bregler}]{dufour2024ammeba}
Nicholas Dufour, Arkanath Pathak, Pouya Samangouei, Nikki Hariri, Shashi
  Deshetti, Andrew Dudfield, Christopher Guess, Pablo~Hern{\'{a}}ndez Escayola,
  Bobby Tran, Mevan Babakar, and Christoph Bregler. 2024.
\newblock \href {https://doi.org/10.48550/ARXIV.2405.11697} {{A}{M}{M}e{B}a:
  {A} {L}arge-scale {S}urvey and {D}ataset of {M}edia-based {M}isinformation
  {I}n-{T}he-wild}.
\newblock \emph{CoRR}, abs/2405.11697.

\bibitem[{Ecker et~al.(2024)Ecker, Tay, Roozenbeek, van~der Linden, Cook,
  Oreskes, and Lewandowsky}]{Ecker2024}
Ullrich K.~H. Ecker, Lena~Q. Tay, Jon Roozenbeek, Sander van~der Linden, John
  Cook, Naomi Oreskes, and Stephan Lewandowsky. 2024.
\newblock \href {https://doi.org/10.1037/amp0001448} {{W}hy misinformation must
  not be ignored}.
\newblock \emph{American Psychologist}.
\newblock Advance online publication.

\bibitem[{Ecker et~al.(2022)Ecker, Lewandowsky, Cook, Schmid, Fazio, Brashier,
  Kendeou, Vraga, and Amazeen}]{ecker2022psychological}
Ullrich~KH Ecker, Stephan Lewandowsky, John Cook, Philipp Schmid, Lisa~K Fazio,
  Nadia Brashier, Panayiota Kendeou, Emily~K Vraga, and Michelle~A Amazeen.
  2022.
\newblock {T}he psychological drivers of misinformation belief and its
  resistance to correction.
\newblock \emph{Nature Reviews Psychology}, 1(1):13--29.

\bibitem[{Geirhos et~al.(2020)Geirhos, Jacobsen, Michaelis, Zemel, Brendel,
  Bethge, and Wichmann}]{geirhos2020shortcut}
Robert Geirhos, J{\"o}rn-Henrik Jacobsen, Claudio Michaelis, Richard Zemel,
  Wieland Brendel, Matthias Bethge, and Felix~A Wichmann. 2020.
\newblock Shortcut learning in deep neural networks.
\newblock \emph{Nature Machine Intelligence}, 2(11):665--673.

\bibitem[{Geng et~al.(2024)Geng, Kementchedjhieva, Nakov, and
  Gurevych}]{geng2024multimodal}
Jiahui Geng, Yova Kementchedjhieva, Preslav Nakov, and Iryna Gurevych. 2024.
\newblock \href {https://doi.org/10.48550/arXiv.2403.03627} {Multimodal large
  language models to support real-world fact-checking}.
\newblock \emph{CoRR}, abs/2403.03627.

\bibitem[{Glockner et~al.(2022)Glockner, Hou, and
  Gurevych}]{glockner2022missing}
Max Glockner, Yufang Hou, and Iryna Gurevych. 2022.
\newblock \href {https://doi.org/10.18653/v1/2022.emnlp-main.397} {Missing
  counter-evidence renders {NLP} fact-checking unrealistic for misinformation}.
\newblock In \emph{Proceedings of the 2022 Conference on Empirical Methods in
  Natural Language Processing}, pages 5916--5936, Abu Dhabi, United Arab
  Emirates. Association for Computational Linguistics.

\bibitem[{Graves(2017)}]{graves2017anatomy}
Lucas Graves. 2017.
\newblock {A}natomy of a fact check: {O}bjective practice and the contested
  epistemology of fact checking.
\newblock \emph{Communication, culture \& critique}, 10(3):518--537.

\bibitem[{Greenspan and Loftus(2021)}]{greenspan2021pandemics}
Rachel~Leigh Greenspan and Elizabeth~F Loftus. 2021.
\newblock {P}andemics and infodemics: {R}esearch on the effects of
  misinformation on memory.
\newblock \emph{Human Behavior and Emerging Technologies}, 3(1):8--12.

\bibitem[{Gurari et~al.(2018)Gurari, Li, Stangl, Guo, Lin, Grauman, Luo, and
  Bigham}]{gurari2018vizwiz}
Danna Gurari, Qing Li, Abigale~J Stangl, Anhong Guo, Chi Lin, Kristen Grauman,
  Jiebo Luo, and Jeffrey~P Bigham. 2018.
\newblock Vizwiz grand challenge: Answering visual questions from blind people.
\newblock In \emph{Proceedings of the IEEE conference on computer vision and
  pattern recognition}, pages 3608--3617.

\bibitem[{Hassid et~al.(2024)Hassid, Remez, Gehring, Schwartz, and
  Adi}]{hassid2024larger}
Michael Hassid, Tal Remez, Jonas Gehring, Roy Schwartz, and Yossi Adi. 2024.
\newblock \href {https://openreview.net/forum?id=QJvfpWSpWm} {The larger the
  better? improved {LLM} code-generation via budget reallocation}.
\newblock In \emph{First Conference on Language Modeling}.

\bibitem[{Hodosh et~al.(2013)Hodosh, Young, and
  Hockenmaier}]{hodosh2013framing}
Micah Hodosh, Peter Young, and Julia Hockenmaier. 2013.
\newblock \href {https://doi.org/10.1613/JAIR.3994} {{F}raming {I}mage
  {D}escription as a {R}anking {T}ask: {D}ata, {M}odels and {E}valuation
  {M}etrics}.
\newblock \emph{J. Artif. Intell. Res.}, 47:853--899.

\bibitem[{Jiang and Wilson(2018)}]{jiang2018linguistic}
Shan Jiang and Christo Wilson. 2018.
\newblock \href {https://doi.org/10.1145/3274351} {{L}inguistic {S}ignals under
  {M}isinformation and {F}act-{C}hecking: {E}vidence from {U}ser {C}omments on
  {S}ocial {M}edia}.
\newblock \emph{Proc. {ACM} Hum. Comput. Interact.}, 2({CSCW}):82:1--82:23.

\bibitem[{Jiao et~al.(2020)Jiao, Yin, Shang, Jiang, Chen, Li, Wang, and
  Liu}]{tinybert}
Xiaoqi Jiao, Yichun Yin, Lifeng Shang, Xin Jiang, Xiao Chen, Linlin Li, Fang
  Wang, and Qun Liu. 2020.
\newblock \href {https://doi.org/10.18653/V1/2020.FINDINGS-EMNLP.372}
  {Tinybert: Distilling {BERT} for natural language understanding}.
\newblock In \emph{Findings of the Association for Computational Linguistics:
  {EMNLP} 2020, Online Event, 16-20 November 2020}, volume {EMNLP} 2020 of
  \emph{Findings of {ACL}}, pages 4163--4174. Association for Computational
  Linguistics.

\bibitem[{Khurana et~al.(2024)Khurana, Nalisnick, Fokkens, and
  Swayamdipta}]{khurana2024crowd}
Urja Khurana, Eric Nalisnick, Antske Fokkens, and Swabha Swayamdipta. 2024.
\newblock \href {https://openreview.net/forum?id=VWWzO3ewMS} {Crowd-calibrator:
  Can annotator disagreement inform calibration in subjective tasks?}
\newblock In \emph{First Conference on Language Modeling}.

\bibitem[{Konstantinovskiy et~al.(2021)Konstantinovskiy, Price, Babakar, and
  Zubiaga}]{konstantinovskiy2021toward}
Lev Konstantinovskiy, Oliver Price, Mevan Babakar, and Arkaitz Zubiaga. 2021.
\newblock \href {https://doi.org/10.1145/3412869} {{T}oward {A}utomated
  {F}actchecking: {D}eveloping an {A}nnotation {S}chema and {B}enchmark for
  {C}onsistent {A}utomated {C}laim {D}etection}.
\newblock \emph{{DTRAP}}, 2(2):14:1--14:16.

\bibitem[{Korshunov and Marcel(2019)}]{Korshunov_ICB_2019}
Pavel Korshunov and S{\'{e}}bastien Marcel. 2019.
\newblock Vulnerability assessment and detection of deepfake videos.
\newblock In \emph{IAPR International Conference on Biometrics},
  Idiap-RR-18-2018.

\bibitem[{Kreps et~al.(2022)Kreps, McCain, and Brundage}]{kreps2022all}
Sarah Kreps, R~Miles McCain, and Miles Brundage. 2022.
\newblock {A}ll the news that’s fit to fabricate: {A}{I}-generated text as a
  tool of media misinformation.
\newblock \emph{Journal of experimental political science}, 9(1):104--117.

\bibitem[{Lasser et~al.(2023)Lasser, Aroyehun, Carrella, Simchon, Garcia, and
  Lewandowsky}]{lasser2023alternative}
Jana Lasser, Segun~T Aroyehun, Fabio Carrella, Almog Simchon, David Garcia, and
  Stephan Lewandowsky. 2023.
\newblock {F}rom alternative conceptions of honesty to alternative facts in
  communications by {U}{S} politicians.
\newblock \emph{Nature human behaviour}, 7(12):2140--2151.

\bibitem[{Li et~al.(2024)Li, Zhang, Zhang, Guo, Zhang, Li, Zhang, Liu, and
  Li}]{li2024llavanext}
Bo~Li, Kaichen Zhang, Hao Zhang, Dong Guo, Renrui Zhang, Feng Li, Yuanhan
  Zhang, Ziwei Liu, and Chunyuan Li. 2024.
\newblock {LLaVA-NeXT: {I}mproved reasoning, {O}{C}{R}, and world knowledge}.
\newblock \url{https://llava-vl.github.io/blog/2024-01-30-llava-next/}.
\newblock Accessed January 16, 2025.

\bibitem[{Li et~al.(2022)Li, Li, Xiong, and Hoi}]{li2022blip}
Junnan Li, Dongxu Li, Caiming Xiong, and Steven Hoi. 2022.
\newblock \href {https://proceedings.mlr.press/v162/li22n.html} {{BLIP}:
  Bootstrapping language-image pre-training for unified vision-language
  understanding and generation}.
\newblock In \emph{Proceedings of the 39th International Conference on Machine
  Learning}, volume 162 of \emph{Proceedings of Machine Learning Research},
  pages 12888--12900. PMLR.

\bibitem[{Lin et~al.(2024)Lin, Gupta, Zhang, Ren, Liu, Ding, Wang, Li,
  Verdoliva, and Hu}]{lin2024detecting}
Li~Lin, Neeraj Gupta, Yue Zhang, Hainan Ren, Chun{-}Hao Liu, Feng Ding, Xin
  Wang, Xin Li, Luisa Verdoliva, and Shu Hu. 2024.
\newblock \href {https://doi.org/10.48550/ARXIV.2402.00045} {{D}etecting
  {M}ultimedia {G}enerated by {L}arge {AI} {M}odels: {A} {S}urvey}.
\newblock \emph{CoRR}, abs/2402.00045.

\bibitem[{Liu et~al.(2023)Liu, Li, Wu, and Lee}]{liu2024visual}
Haotian Liu, Chunyuan Li, Qingyang Wu, and Yong~Jae Lee. 2023.
\newblock \href
  {https://proceedings.neurips.cc/paper_files/paper/2023/file/6dcf277ea32ce3288914faf369fe6de0-Paper-Conference.pdf}
  {Visual instruction tuning}.
\newblock In \emph{Advances in Neural Information Processing Systems},
  volume~36, pages 34892--34916. Curran Associates, Inc.

\bibitem[{Liu et~al.(2024)Liu, Li, Li, Xia, Cui, Huang, Huang, Deng, and
  He}]{liu2024mmfakebench}
Xuannan Liu, Zekun Li, Peipei Li, Shuhan Xia, Xing Cui, Linzhi Huang, Huaibo
  Huang, Weihong Deng, and Zhaofeng He. 2024.
\newblock \href {https://doi.org/10.48550/ARXIV.2406.08772}
  {{M}{M}{F}ake{B}ench: {A} {M}ixed-source {M}ultimodal {M}isinformation
  {D}etection {B}enchmark for {L}{V}{L}{M}s}.
\newblock \emph{CoRR}, abs/2406.08772.

\bibitem[{Luo et~al.(2021)Luo, Darrell, and Rohrbach}]{luo2021newsclippings}
Grace Luo, Trevor Darrell, and Anna Rohrbach. 2021.
\newblock \href {https://doi.org/10.18653/v1/2021.emnlp-main.545}
  {{N}ews{CLIP}pings: {A}utomatic {G}eneration of {O}ut-of-context {M}ultimodal
  {M}edia}.
\newblock In \emph{Proceedings of the 2021 Conference on Empirical Methods in
  Natural Language Processing}, pages 6801--6817, Online and Punta Cana,
  Dominican Republic. Association for Computational Linguistics.

\bibitem[{Madras et~al.(2018)Madras, Pitassi, and Zemel}]{madras2018predict}
David Madras, Toniann Pitassi, and Richard~S. Zemel. 2018.
\newblock \href
  {https://proceedings.neurips.cc/paper/2018/hash/09d37c08f7b129e96277388757530c72-Abstract.html}
  {{P}redict {R}esponsibly: {I}mproving {F}airness and {A}ccuracy by {L}earning
  to {D}efer}.
\newblock In \emph{Advances in Neural Information Processing Systems 31: Annual
  Conference on Neural Information Processing Systems 2018, NeurIPS 2018,
  December 3-8, 2018, Montr{\'{e}}al, Canada}, pages 6150--6160.

\bibitem[{Majer and Snajder(2024)}]{majer2024claim}
Laura Majer and Jan Snajder. 2024.
\newblock \href {https://doi.org/10.48550/ARXIV.2404.12174} {{C}laim
  {C}heck-worthiness {D}etection: {H}ow {W}ell do {L}{L}{M}s {G}rasp
  {A}nnotation {G}uidelines?}
\newblock \emph{CoRR}, abs/2404.12174.

\bibitem[{McCright and Dunlap(2017)}]{mccright2017combatting}
Aaron~M McCright and Riley~E Dunlap. 2017.
\newblock {C}ombatting misinformation requires recognizing its types and the
  factors that facilitate its spread and resonance.
\newblock \emph{Journal of Applied Research in Memory and Cognition}.

\bibitem[{Micallef et~al.(2022)Micallef, Armacost, Memon, and
  Patil}]{micallef2022true}
Nicholas Micallef, Vivienne Armacost, Nasir~D. Memon, and Sameer Patil. 2022.
\newblock \href {https://doi.org/10.1145/3512974} {{T}rue or {F}alse:
  {S}tudying the {W}ork {P}ractices of {P}rofessional {F}act-checkers}.
\newblock \emph{Proc. {ACM} Hum. Comput. Interact.}, 6({CSCW1}):127:1--127:44.

\bibitem[{{Midjourney, Inc.}(2023)}]{midjourney2023midjourney}
{Midjourney, Inc.} 2023.
\newblock \href {https://www.midjourney.com/} {{Midjourney}}.
\newblock Generative AI model for image generation (Version 6).

\bibitem[{{Mistral}(2024)}]{mistral2024pixtral}
{Mistral}. 2024.
\newblock \href {https://mistral.ai/news/pixtral-12b/} {{A}nnouncing {P}ixtral
  12{B}}.

\bibitem[{Monteith et~al.(2024)Monteith, Glenn, Geddes, Whybrow, Achtyes, and
  Bauer}]{monteith2024artificial}
Scott Monteith, Tasha Glenn, John~R Geddes, Peter~C Whybrow, Eric Achtyes, and
  Michael Bauer. 2024.
\newblock \href
  {https://www.cambridge.org/core/journals/the-british-journal-of-psychiatry/article/artificial-intelligence-and-increasing-misinformation/DCCE0EB214E3D375A3006AA69FFB210D}
  {Artificial intelligence and increasing misinformation}.
\newblock \emph{The British Journal of Psychiatry}, 224(2):33--35.

\bibitem[{Nakamura et~al.(2020)Nakamura, Levy, and Wang}]{nakamura2020fakeddit}
Kai Nakamura, Sharon Levy, and William~Yang Wang. 2020.
\newblock \href {https://aclanthology.org/2020.lrec-1.755/} {{F}akeddit: {A}
  {N}ew {M}ultimodal {B}enchmark {D}ataset for {F}ine-grained {F}ake {N}ews
  {D}etection}.
\newblock In \emph{Proceedings of The 12th Language Resources and Evaluation
  Conference, {LREC} 2020, Marseille, France, May 11-16, 2020}, pages
  6149--6157. European Language Resources Association.

\bibitem[{Nakov et~al.(2018)Nakov, Barr{\'o}n-Cedeno, Elsayed, Suwaileh,
  M{\`a}rquez, Zaghouani, Atanasova, Kyuchukov, and
  Da~San~Martino}]{nakov2018overview}
Preslav Nakov, Alberto Barr{\'o}n-Cedeno, Tamer Elsayed, Reem Suwaileh,
  Llu{\'\i}s M{\`a}rquez, Wajdi Zaghouani, Pepa Atanasova, Spas Kyuchukov, and
  Giovanni Da~San~Martino. 2018.
\newblock Overview of the clef-2018 checkthat! lab on automatic identification
  and verification of political claims.
\newblock In \emph{Experimental IR Meets Multilinguality, Multimodality, and
  Interaction: 9th International Conference of the CLEF Association, CLEF 2018,
  Avignon, France, September 10-14, 2018, Proceedings 9}, pages 372--387.
  Springer.

\bibitem[{Nakov et~al.(2021)Nakov, Corney, Hasanain, Alam, Elsayed,
  Barr{\'o}n-Cede{\~n}o, Papotti, Shaar, Da~San~Martino
  et~al.}]{nakov2021automated}
Preslav Nakov, David Corney, Maram Hasanain, Firoj Alam, Tamer Elsayed, Alberto
  Barr{\'o}n-Cede{\~n}o, Paolo Papotti, Shaden Shaar, Giovanni Da~San~Martino,
  et~al. 2021.
\newblock Automated fact-checking for assisting human fact-checkers.
\newblock In \emph{Proceedings of the Thirtieth International Joint Conference
  onArtificial Intelligence,$\{$IJCAI-21$\}$}, pages 4551--4558. International
  Joint Conferences on Artificial Intelligence Organization.

\bibitem[{OpenAI(2023)}]{openai2023chatgpt}
OpenAI. 2023.
\newblock \href {https://chat.openai.com/} {{ChatGPT}}.
\newblock Large language model.

\bibitem[{Pan et~al.(2023)Pan, Pan, Chen, Nakov, Kan, and Wang}]{pan2023risk}
Yikang Pan, Liangming Pan, Wenhu Chen, Preslav Nakov, Min-Yen Kan, and William
  Wang. 2023.
\newblock \href {https://doi.org/10.18653/v1/2023.findings-emnlp.97} {On the
  risk of misinformation pollution with large language models}.
\newblock In \emph{Findings of the Association for Computational Linguistics:
  EMNLP 2023}, pages 1389--1403, Singapore. Association for Computational
  Linguistics.

\bibitem[{Papadopoulos et~al.(2023)Papadopoulos, Koutlis, Papadopoulos, and
  Petrantonakis}]{papadopoulos2023synthetic}
Stefanos-Iordanis Papadopoulos, Christos Koutlis, Symeon Papadopoulos, and
  Panagiotis Petrantonakis. 2023.
\newblock \href {https://doi.org/10.1145/3592572.3592842} {Synthetic
  misinformers: Generating and combating multimodal misinformation}.
\newblock In \emph{Proceedings of the 2nd ACM International Workshop on
  Multimedia AI against Disinformation}, MAD '23, page 36–44, New York, NY,
  USA. Association for Computing Machinery.

\bibitem[{Papadopoulos et~al.(2024)Papadopoulos, Koutlis, Papadopoulos, and
  Petrantonakis}]{papadopoulos2024verite}
Stefanos{-}Iordanis Papadopoulos, Christos Koutlis, Symeon Papadopoulos, and
  Panagiotis~C. Petrantonakis. 2024.
\newblock \href {https://doi.org/10.1007/S13735-023-00312-6} {{VERITE:} a
  {R}obust benchmark for multimodal misinformation detection accounting for
  unimodal bias}.
\newblock \emph{Int. J. Multim. Inf. Retr.}, 13(1):4.

\bibitem[{Perez and Ribeiro(2022)}]{perez2022ignore}
F{\'a}bio Perez and Ian Ribeiro. 2022.
\newblock \href {https://openreview.net/forum?id=qiaRo_7Zmug} {Ignore previous
  prompt: Attack techniques for language models}.
\newblock In \emph{NeurIPS ML Safety Workshop}.

\bibitem[{Pikuliak et~al.(2023)Pikuliak, Srba, Moro, Hromadka, Smole{\v{n}},
  Meli{\v{s}}ek, Vykopal, Simko, Podrou{\v{z}}ek, and
  Bielikova}]{pikuliak2023multilingual}
Mat{\'u}{\v{s}} Pikuliak, Ivan Srba, Robert Moro, Timo Hromadka, Timotej
  Smole{\v{n}}, Martin Meli{\v{s}}ek, Ivan Vykopal, Jakub Simko, Juraj
  Podrou{\v{z}}ek, and Maria Bielikova. 2023.
\newblock \href {https://doi.org/10.18653/v1/2023.emnlp-main.1027}
  {Multilingual previously fact-checked claim retrieval}.
\newblock In \emph{Proceedings of the 2023 Conference on Empirical Methods in
  Natural Language Processing}, pages 16477--16500, Singapore. Association for
  Computational Linguistics.

\bibitem[{Procter et~al.(2023)Procter, Arana{-}Catania, He, Liakata, Zubiaga,
  Kochkina, and Zhao}]{procter2023some}
Rob Procter, Miguel Arana{-}Catania, Yulan He, Maria Liakata, Arkaitz Zubiaga,
  Elena Kochkina, and Runcong Zhao. 2023.
\newblock \href {https://doi.org/10.48550/ARXIV.2305.02224} {{S}ome
  {O}bservations on {F}act-checking {W}ork with {I}mplications for
  {C}omputational {S}upport}.
\newblock \emph{CoRR}, abs/2305.02224.

\bibitem[{Ribeiro et~al.(2017)Ribeiro, Calais, Almeida, and
  Jr.}]{ribeiro2017everything}
Manoel~Horta Ribeiro, Pedro~H. Calais, Virg{\'{\i}}lio A.~F. Almeida, and
  Wagner~Meira Jr. 2017.
\newblock \href {https://arxiv.org/abs/1706.05924} {"{E}verything {I}
  {D}isagree {W}ith is {\#}fakenews": {C}orrelating {P}olitical {P}olarization
  and {S}pread of {M}isinformation}.
\newblock \emph{CoRR}, abs/1706.05924.

\bibitem[{Rocha et~al.(2021)Rocha, De~Moura, Desid{\'e}rio, De~Oliveira,
  Louren{\c{c}}o, and de~Figueiredo~Nicolete}]{rocha2021impact}
Yasmim~Mendes Rocha, Gabriel~Ac{\'a}cio De~Moura, Gabriel~Alves Desid{\'e}rio,
  Carlos~Henrique De~Oliveira, Francisco~Dantas Louren{\c{c}}o, and
  Larissa~Deadame de~Figueiredo~Nicolete. 2021.
\newblock {T}he impact of fake news on social media and its influence on health
  during the {C}{O}{V}{I}{D}-19 pandemic: {A} systematic review.
\newblock \emph{Journal of Public Health}, pages 1--10.

\bibitem[{Schlichtkrull et~al.(2023)Schlichtkrull, Guo, and
  Vlachos}]{schlichtkrull2024averitec}
Michael Schlichtkrull, Zhijiang Guo, and Andreas Vlachos. 2023.
\newblock \href
  {https://proceedings.neurips.cc/paper_files/paper/2023/file/cd86a30526cd1aff61d6f89f107634e4-Paper-Datasets_and_Benchmarks.pdf}
  {Averitec: A dataset for real-world claim verification with evidence from the
  web}.
\newblock In \emph{Advances in Neural Information Processing Systems},
  volume~36, pages 65128--65167. Curran Associates, Inc.

\bibitem[{Seow et~al.(2022)Seow, Lim, Phan, and Liu}]{seow2022comprehensive}
Jia~Wen Seow, Mei~Kuan Lim, Rapha{\"{e}}l~C.{-}W. Phan, and Joseph~K. Liu.
  2022.
\newblock \href {https://doi.org/10.1016/J.NEUCOM.2022.09.135} {{A}
  comprehensive overview of {D}eepfake: {G}eneration, detection, datasets, and
  opportunities}.
\newblock \emph{Neurocomputing}, 513:351--371.

\bibitem[{Sharma et~al.(2019)Sharma, Qian, Jiang, Ruchansky, Zhang, and
  Liu}]{sharma2019combating}
Karishma Sharma, Feng Qian, He~Jiang, Natali Ruchansky, Ming Zhang, and Yan
  Liu. 2019.
\newblock \href {https://doi.org/10.1145/3305260} {Combating fake news: A
  survey on identification and mitigation techniques}.
\newblock \emph{ACM Trans. Intell. Syst. Technol.}, 10(3).

\bibitem[{Sharma et~al.(2018)Sharma, Ding, Goodman, and
  Soricut}]{sharma2018conceptual}
Piyush Sharma, Nan Ding, Sebastian Goodman, and Radu Soricut. 2018.
\newblock \href {https://doi.org/10.18653/V1/P18-1238} {{C}onceptual
  {C}aptions: {A} {C}leaned, {H}ypernymed, {I}mage {A}lt-text {D}ataset {F}or
  {A}utomatic {I}mage {C}aptioning}.
\newblock In \emph{Proceedings of the 56th Annual Meeting of the Association
  for Computational Linguistics, {ACL} 2018, Melbourne, Australia, July 15-20,
  2018, Volume 1: Long Papers}, pages 2556--2565. Association for Computational
  Linguistics.

\bibitem[{Singal et~al.(2024)Singal, Patwa, Patwa, Chadha, and
  Das}]{singal2024evidence}
Ronit Singal, Pransh Patwa, Parth Patwa, Aman Chadha, and Amitava Das. 2024.
\newblock \href {https://doi.org/10.18653/v1/2024.fever-1.10}
  {{E}vidence-backed {F}act {C}hecking using {RAG} and {F}ew-shot {I}n-context
  {L}earning with {LLM}s}.
\newblock In \emph{Proceedings of the Seventh Fact Extraction and VERification
  Workshop (FEVER)}, pages 91--98, Miami, Florida, USA. Association for
  Computational Linguistics.

\bibitem[{Singh and Sharma(2022)}]{singh2022predicting}
Bhuvanesh Singh and Dilip~Kumar Sharma. 2022.
\newblock \href {https://doi.org/10.1007/S00521-021-06086-4} {{P}redicting
  image credibility in fake news over social media using multi-modal approach}.
\newblock \emph{Neural Comput. Appl.}, 34(24):21503--21517.

\bibitem[{Singla et~al.(2024)Singla, Yue, Paul, Shirkavand, Jayawardhana,
  Ganjdanesh, Huang, Bhatele, Somepalli, and Goldstein}]{singla2024pixels}
Vasu Singla, Kaiyu Yue, Sukriti Paul, Reza Shirkavand, Mayuka Jayawardhana,
  Alireza Ganjdanesh, Heng Huang, Abhinav Bhatele, Gowthami Somepalli, and Tom
  Goldstein. 2024.
\newblock \href {https://doi.org/10.48550/ARXIV.2406.10328} {{F}rom {P}ixels to
  {P}rose: {A} {L}arge {D}ataset of {D}ense {I}mage {C}aptions}.
\newblock \emph{CoRR}, abs/2406.10328.

\bibitem[{Soprano et~al.(2024)Soprano, Roitero, Barbera, Ceolin, Spina,
  Demartini, and Mizzaro}]{soprano2024cognitive}
Michael Soprano, Kevin Roitero, David~La Barbera, Davide Ceolin, Damiano Spina,
  Gianluca Demartini, and Stefano Mizzaro. 2024.
\newblock \href {https://doi.org/10.1016/J.IPM.2024.103672} {{C}ognitive
  {B}iases in {F}act-checking and {T}heir {C}ountermeasures: {A} {R}eview}.
\newblock \emph{Inf. Process. Manag.}, 61(2):103672.

\bibitem[{Stability.ai(2024)}]{stablediff}
Stability.ai. 2024.
\newblock \href {https://stability.ai/news/introducing-stable-diffusion-3-5}
  {{S}table {D}iffusion 3.5}.

\bibitem[{Stepanova and Ross(2023)}]{stepanova-ross-2023-temporal}
Nataliya Stepanova and Bj{\"o}rn Ross. 2023.
\newblock \href {https://doi.org/10.18653/v1/2023.genbench-1.6} {{T}emporal
  {G}eneralizability in {M}ultimodal {M}isinformation {D}etection}.
\newblock In \emph{Proceedings of the 1st GenBench Workshop on (Benchmarking)
  Generalisation in NLP}, pages 76--88, Singapore. Association for
  Computational Linguistics.

\bibitem[{Tonglet et~al.(2024)Tonglet, Moens, and Gurevych}]{tonglet2024image}
Jonathan Tonglet, Marie-Francine Moens, and Iryna Gurevych. 2024.
\newblock \href {https://doi.org/10.18653/v1/2024.emnlp-main.448}
  {{\textquotedblleft}image, tell me your story!{\textquotedblright} predicting
  the original meta-context of visual misinformation}.
\newblock In \emph{Proceedings of the 2024 Conference on Empirical Methods in
  Natural Language Processing}, pages 7845--7864, Miami, Florida, USA.
  Association for Computational Linguistics.

\bibitem[{van~der Meer(2024)}]{vandermeer2024facilitating}
Michiel van~der Meer. 2024.
\newblock \href {https://doi.org/10.18653/v1/2024.naacl-srw.29} {{F}acilitating
  {O}pinion {D}iversity through {H}ybrid {NLP} {A}pproaches}.
\newblock In \emph{Proceedings of the 2024 Conference of the North American
  Chapter of the Association for Computational Linguistics: Human Language
  Technologies (Volume 4: Student Research Workshop)}, pages 272--284, Mexico
  City, Mexico. Association for Computational Linguistics.

\bibitem[{van~der Meer et~al.(2024{\natexlab{a}})van~der Meer, Falk,
  Murukannaiah, and Liscio}]{vandermeer2024annotator}
Michiel van~der Meer, Neele Falk, Pradeep~K. Murukannaiah, and Enrico Liscio.
  2024{\natexlab{a}}.
\newblock \href {https://doi.org/10.18653/v1/2024.emnlp-main.1031}
  {Annotator-centric active learning for subjective {NLP} tasks}.
\newblock In \emph{Proceedings of the 2024 Conference on Empirical Methods in
  Natural Language Processing}, pages 18537--18555, Miami, Florida, USA.
  Association for Computational Linguistics.

\bibitem[{van~der Meer et~al.(2024{\natexlab{b}})van~der Meer, Liscio, Jonker,
  Plaat, Vossen, and Murukannaiah}]{vandermeer2024hybrid}
Michiel van~der Meer, Enrico Liscio, Catholijn Jonker, Aske Plaat, Piek Vossen,
  and Pradeep Murukannaiah. 2024{\natexlab{b}}.
\newblock A hybrid intelligence method for argument mining.
\newblock \emph{Journal of Artificial Intelligence Research}, 80:1187--1222.

\bibitem[{Vogel and Frick(2023)}]{frick2023fraunhofer}
Inna Vogel and Raphael Frick. 2023.
\newblock \href {https://doi.org/10.48550/arXiv.2307.00610} {Fraunhofer sit at
  checkthat! 2023: Mixing single-modal classifiers to estimate the
  check-worthiness of multi-modal tweets}.

\bibitem[{Wolf et~al.(2019)Wolf, Debut, Sanh, Chaumond, Delangue, Moi, Cistac,
  Rault, Louf, Funtowicz, and Brew}]{wolf2019huggingface}
Thomas Wolf, Lysandre Debut, Victor Sanh, Julien Chaumond, Clement Delangue,
  Anthony Moi, Pierric Cistac, Tim Rault, R{\'{e}}mi Louf, Morgan Funtowicz,
  and Jamie Brew. 2019.
\newblock \href {https://arxiv.org/abs/1910.03771} {{H}ugging{F}ace's
  {T}ransformers: {S}tate-of-the-art {N}atural {L}anguage {P}rocessing}.
\newblock \emph{CoRR}, abs/1910.03771.

\bibitem[{Xu et~al.(2023)Xu, Fan, and Kankanhalli}]{xu2023combating}
Danni Xu, Shaojing Fan, and Mohan~S. Kankanhalli. 2023.
\newblock \href {https://doi.org/10.1145/3581783.3612704} {{C}ombating
  {M}isinformation in the {E}ra of {G}enerative {AI} {M}odels}.
\newblock In \emph{Proceedings of the 31st {ACM} International Conference on
  Multimedia, {MM} 2023, Ottawa, ON, Canada, 29 October 2023- 3 November 2023},
  pages 9291--9298. {ACM}.

\bibitem[{Ye(2023)}]{calflops}
Xiaoju Ye. 2023.
\newblock \href {https://github.com/MrYxJ/calculate-flops.pytorch} {calflops: a
  {F}{L}{O}{P}s and {P}arams calculate tool for neural networks in pytorch
  framework}.

\bibitem[{Yoon et~al.(2024)Yoon, Yoon, and Park}]{yoon2024assessing}
Yejun Yoon, Seunghyun Yoon, and Kunwoo Park. 2024.
\newblock \href {https://doi.org/10.18653/V1/2024.FINDINGS-ACL.534}
  {{A}ssessing {N}ews {T}humbnail {R}epresentativeness: {C}ounterfactual text
  can enhance the cross-modal matching ability}.
\newblock In \emph{Findings of the Association for Computational Linguistics,
  {ACL} 2024, Bangkok, Thailand and virtual meeting, August 11-16, 2024}, pages
  9009--9024. Association for Computational Linguistics.

\bibitem[{Zhou et~al.(2023)Zhou, Zhang, Luo, Parker, and
  Choudhury}]{zhou2023synthetic}
Jiawei Zhou, Yixuan Zhang, Qianni Luo, Andrea~G. Parker, and Munmun~De
  Choudhury. 2023.
\newblock \href {https://doi.org/10.1145/3544548.3581318} {{S}ynthetic {L}ies:
  {U}nderstanding {A}{I}-generated {M}isinformation and {E}valuating
  {A}lgorithmic and {H}uman {S}olutions}.
\newblock In \emph{Proceedings of the 2023 {CHI} Conference on Human Factors in
  Computing Systems, {CHI} 2023, Hamburg, Germany, April 23-28, 2023}, pages
  436:1--436:20. {ACM}.

\bibitem[{Zlatkova et~al.(2019)Zlatkova, Nakov, and Koychev}]{zlatkova2019fact}
Dimitrina Zlatkova, Preslav Nakov, and Ivan Koychev. 2019.
\newblock \href {https://doi.org/10.18653/v1/D19-1216} {Fact-checking meets
  fauxtography: Verifying claims about images}.
\newblock In \emph{Proceedings of the 2019 Conference on Empirical Methods in
  Natural Language Processing and the 9th International Joint Conference on
  Natural Language Processing (EMNLP-IJCNLP)}, pages 2099--2108, Hong Kong,
  China. Association for Computational Linguistics.

\bibitem[{Zou et~al.(2016)Zou, Xie, Lin, Wu, and Ju}]{ZOU20162}
Quan Zou, Sifa Xie, Ziyu Lin, Meihong Wu, and Ying Ju. 2016.
\newblock \href {https://doi.org/10.1016/j.bdr.2015.12.001} {Finding the best
  classification threshold in imbalanced classification}.
\newblock \emph{Big Data Research}, 5:2--8.
\newblock Big data analytics and applications.

\end{thebibliography}


\appendix

\section{Data}
\subsection{Multiclaim}
\label{app:multiclaim-extraction}
To extract images and claims from the Multiclaim dataset, we followed these steps:
\begin{enumerate}
    \item Obtain fact-checking article from URL.
    \item Filter claims for multimodal terms (e.g. ``photo'', ``image'', etc.).
    \item Filter out articles not written in English.
    \item Obtain the image associated with the claim based on the HTML in the article. We look for the image tag that is closest to the claim in the HTML tree.
    \item Filter out some erroneously obtained images based on their URL, such as repeated entries (usually website logos), or specific image dimensions (image too small or aspect ratio too distorted).
\end{enumerate}

\subsection{Multimodal Terms}
\label{app:multimodal-terms}
The complete list of multimodal terms is: photo, image, picture, screenshot, artwork, video.



\section{Experimental Details}
\label{app:experimental-details}

\paragraph{Computational resources}
Experiments were largely run between August 2024 and February 2025. Training and inference were performed on a cluster with heterogeneous computing infrastructure, including RTX3090, V100, and H100 GPUs. We fine-tuned a total of eight models for checkworthiness detection, which took up to two hours per model. For all our experiments, we use the Huggingface transformers library \citep{wolf2019huggingface}, with default values unless otherwise mentioned.

\paragraph{Model versions}
See Table~\ref{tab:model_versions} for the specific checkpoints used to instantiate the FT and ICL models.

\begin{table*}[t]
    \centering
    \begin{tabular}{lll}
         \toprule
         \textbf{Model} & \textbf{Checkpoint} & \textbf{Tuned Treshold}\\
         \midrule
         TinyBERT & \texttt{huawei-noah/TinyBERT\_General\_4L\_312D}    & 0.37249295339780664,\\
         BERT-base & \texttt{google-bert/bert-base-cased}               & 0.000887640770134563\\
         BERT-large & \texttt{google-bert/bert-large-cased}             & 0.0006220573599437401\\
         ResNet & \texttt{microsoft/resnet-26}           2               & 0.4326931064840402\\
         ViT-base & \texttt{google/vit-base-patch16-224-in21k}          & 0.13248605867961713\\
         ViT-large & \texttt{google/vit-large-patch16-224-in21k}        & 0.09581064554051821\\
         BLIP & \texttt{Salesforce/blip-vqa-base}                       & 0.07592173944742421\\
         BLIP2 & \texttt{Salesforce/blip2-itm-vit-g}                    & 0.0761946809023745\\
         Llava & \texttt{llava-hf/llava-v1.6-mistral-7b-hf}             & n/a\\
         Pixtral & \texttt{mistral-community/pixtral-12b}               & n/a\\
         \bottomrule
    \end{tabular}
    \caption{Model checkpoint names used in the experiments as found on the Huggingface Hub.}
    \label{tab:model_versions}
\end{table*}

\subsection{Fine-tuning}
We use hyperparameters shown in Table~\ref{tab:hyperparams} when fine-tuning the models on the training set of CheckThat. Non-mentioned parameters are set using the default values in the Huggingface library. During training, we keep track of the accuracy on the validation set, and at the end of training all epochs, we use the model at the step that obtained the best accuracy on the validation set.

\begin{table}[t]
    \centering
    \begin{tabular}{ll}
         \toprule
         \textbf{Hyperparameter} & \textbf{Value} \\
         \midrule
         learning rate & 2e-05\\
         max epochs & 10\\
         batch size & 16\\
         \bottomrule
    \end{tabular}
    \caption{Fine-tuning hyperparameters}
    \label{tab:hyperparams}
\end{table}

\subsection{In-Context Learning}
For the short prompt, see Prompt~\ref{prompt:short}. For the verbose prompt, see Prompt~\ref{prompt:long}. In cases of few-shot learning where $n>0$, the orange examples are repeated $n$ times, once for each randomly retrieved example. The blue text is filled during inference for each sample in the evaluation set.

\begin{promptbox}[label=prompt:short]{Short prompt}
    The goal of this task is to assess whether a given statement posted by an individual is worth fact-checking. Provide your answer in by selecting between 'checkworthy' or 'not checkworthy', and provide a brief explanation. Give your response in the following format: \{'label': <answer>\}. \\

    {\color{ibm3} Here are some examples: \\
    STATEMENT: <demonstration 1 text>\\
    <demonstration 1 image>\\
    EXAMPLE OUTPUT: <demonstration 1 label>}\\

    STATEMENT: {\color{ibm1} <input text>}\\
    {\color{ibm1} <input image>}\\

    Give your response as a JSON object.
\end{promptbox}

\begin{promptbox}[label=prompt:long]{Verbose prompt}
    The goal of this task is to assess whether a given statement posted by an individual is worth fact-checking. In order to make that decision, one would need to ponder about questions, such as 'does it contain a verifiable factual claim?' or 'is it harmful?', before deciding on the final check-worthiness label. (Multimodality) Given a tweet with the text and its corresponding image, predict whether it is worth fact-checking. Answers to the questions relevant for deriving a label are based on both the image and the text. The image plays two roles for check-worthiness estimation: (i) there is a piece of evidence (e.g., an event, an action, a situation, a person’s identity, etc.) or illustration of certain aspects from the textual claim, and/or (ii) the image contains overlaid text that contains a claim (e.g., misrepresented facts and figures) in a textual form. Provide your answer in by selecting between 'checkworthy' or 'not checkworthy', and providing a brief explanation. Give your response in the following format: \{'label': <answer>\}. \\

    {\color{ibm3} Here are some examples: \\
    STATEMENT: <demonstration 1 text>\\
    <demonstration 1 image>\\
    EXAMPLE OUTPUT: <demonstration 1 label>}\\

    STATEMENT: {\color{ibm1} <input text>}\\
    {\color{ibm1} <input image>}\\

    Give your response as a JSON object.
    \end{promptbox}

\section{Additional Experiments}
\subsection{Threshold tuning for fine-tuned models}
\label{app:tuning-threshold}
\begin{figure*}
    \centering
    \includegraphics[width=\linewidth]{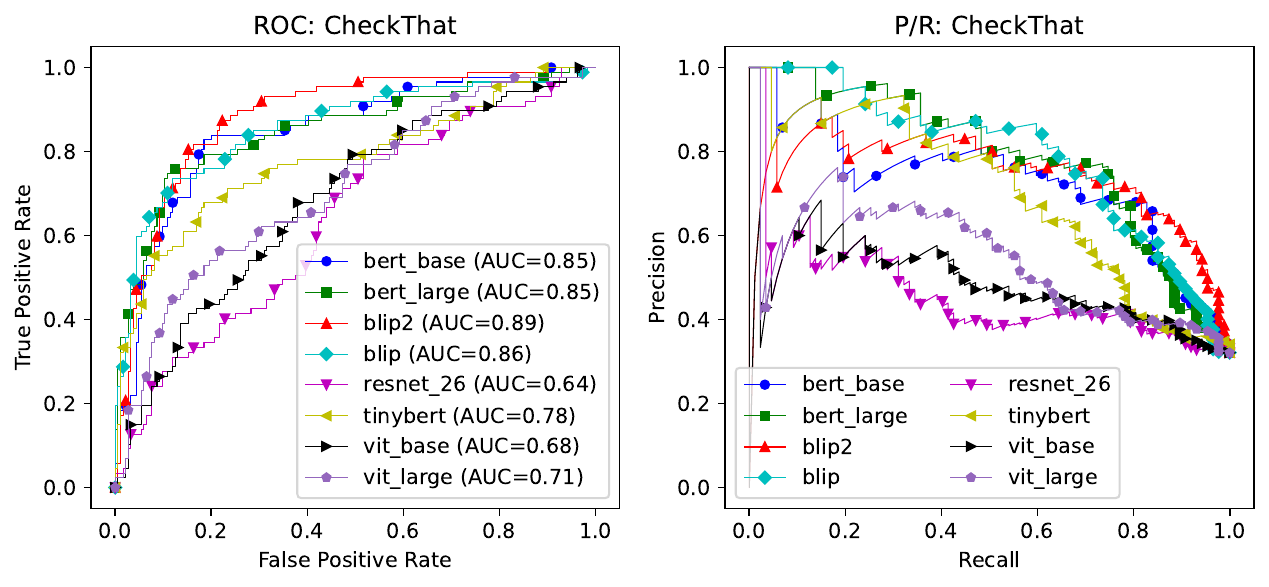}
    \caption{Tuning threshold parameter for CheckThat dataset. (Left) ROC for the fine-tuned models. (Right) Precision/Recall tradeoff.}
    \label{fig:threshold-checkthat}
\end{figure*}

\begin{figure*}
    \centering
    \includegraphics[width=\linewidth]{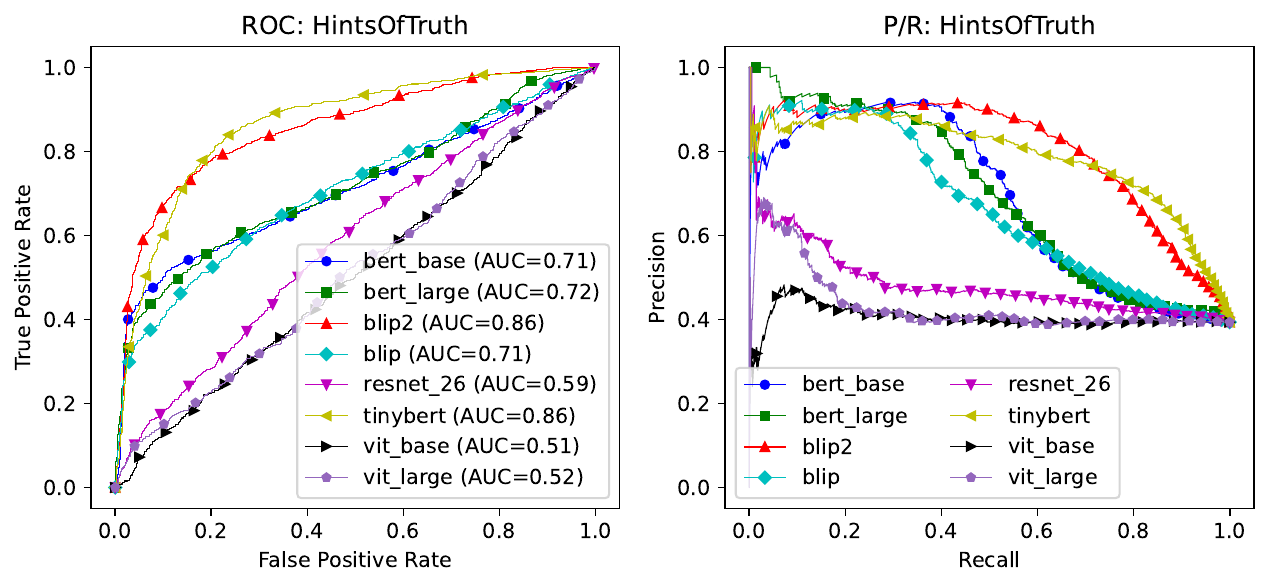}
    \caption{Tuning threshold parameter for HintsOfTruth. (Left) ROC for the fine-tuned models. (Right) Precision/Recall tradeoff.}
    \label{fig:threshold-hints}
\end{figure*}

See Figures~\ref{fig:threshold-checkthat} and~\ref{fig:threshold-hints} for the ROC and Precicion-Recall curves for CheckThat and \hiot, respectively. We selected a False Positive Rate (FPR) of 0.3 as the threshold point to prefer recall over precision. The final threshold values are reported in Table~\ref{tab:model_versions}.

\subsection{Single-logit Output}
\label{app:single-logit}
\begin{figure}
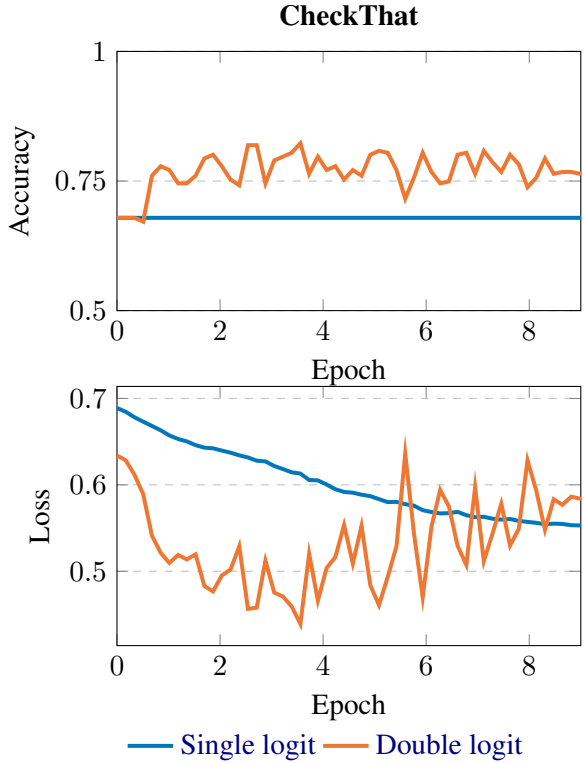

    \centering
    \includestandalone[width=\columnwidth]{plots/app_single_vs_double}
    \caption{Performance during training a TinyBERT model using a single or double logit setup.}
    \label{fig:single-logit-app}
\end{figure}

Opposed to the two-logit setup used in the experiments in this paper, one could use a single-logit setup to perform checkworthiness detection. In this setup, we would turn the softmax and cross-entropy loss into a sigmoid activation and a binary cross-entropy loss. However, we found this led did not let the model learn better-than-random accuracy, even though its loss was going down more smoothly when training on the CheckThat data, see Figure~\ref{fig:single-logit-app}.

\subsection{Image captions as Negative Samples}
\label{app:image-captions}
In this set of experiments, we investigate whether models for multimodal checkworthiness are likely to label generic image captions as checkworthy wrongly. This acts as an additional sanity check that our models do not rely on spurious features or other shortcuts \citep{geirhos2020shortcut}.

\paragraph{Approach} We take models trained on the CheckThat dataset, and apply them across various image caption datasets. Since image captions \begin{enumerate*}[label=(\arabic*)]
    \item do not contain verifiable claims, or
    \item are not (potentially) harmful,
\end{enumerate*} they can be considered non-checkworthy. We perform experiments on the following datasets:
\begin{enumerate*}[label=(\arabic*)]
    \item \textbf{Flickr30K} \citep{hodosh2013framing}
    \item \textbf{VizWiz} \citep{gurari2018vizwiz}
    \item \textbf{Conceptual Captions} \citep{sharma2018conceptual}
    \item \textbf{Coyo} \citep{kakaobrain2022coyo-700m}
    \item \textbf{PixelProse} \citep{singla2024pixels}
\end{enumerate*}. For each dataset, impose similar filtering on the textual claims as for the scraped datasets mentioned in Section~\ref{sec:method}. Furthermore, since these datasets are of significant size, we downsample them to 6K samples each before accessing the image URLs. Table~\ref{app:tab:negative-datasets} denotes the final sizes of each dataset. We further split this into training/test/validation sets using a 70/20/10 ratio. We then apply each checkworthiness detection approach (as described in Section~\ref{sec:experimental-approaches}) to this task and report the classification accuracy.

\begin{table}[t]
    \centering
    \begin{tabular}{lcc}
        \toprule
        \textbf{Dataset} & \textbf{Source}& \textbf{Size} \\
        \midrule
        Flickr30K & Flickr              & 3,000 \\
        VizWiz & Blind humans           & 693 \\
        Conceptual Captions & Webpages  & 3,168\\
        Coyo & Webpages                 & 3,872\\
        PixelProse & Webpages           & 5,331\\
        \bottomrule
    \end{tabular}
    \caption{Additional image captioning datasets besides Flickr30K.}
    \label{app:tab:negative-datasets}
\end{table}

\paragraph{Results}
\begin{table}[t]
    \centering
    \begin{tabular}{lccccc}
        \toprule
        & \multicolumn{5}{c}{Accuracy}\\
        \textbf{Model} & \textbf{Fl} & \textbf{VW} & \textbf{CC} &\textbf{Co} &\textbf{PP}  \\
        \midrule
TinyBERT         & .787 & .775 & .593 & .586 & .360 \\
BERT-base        & .582 & .444 & .695 & .875 & .472 \\
BERT-large       & .472 & .092 & .383 & .775 & .111 \\
ResNet-26        & .713 & .641 & .697 & .671 & .688 \\
ViT-base         & .708 & .754 & .796 & .809 & .800 \\
ViT-large        & .703 & .831 & .790 & .783 & .750 \\
BLIP             & .913 & .317 & .170 & .423 & .065 \\
BLIP2            & .739 & .930 & .830 & .900 & .806 \\
Llava (0)        & .047 & .626 & .563 & .740 & .359 \\
Pixtral (0)      & .923 & .713 & .597 & .363 & .577 \\
        \bottomrule
    \end{tabular}
    \caption{Accuracies for each of the image captioning datasets: Flickr30K (\textbf{Fl}), VizWiz (\textbf{VW}), Conceptual Captions (\textbf{CC}), Coyo (\textbf{Co}), and PixelProse (\textbf{PP}).}
    \label{app:tab:negative-datasets-results}
\end{table}

See Table~\ref{app:tab:negative-datasets-results}.

\subsection{Compute Wall Time}
\label{app:wall-time}
\begin{figure}
    \centering
    \begin{tikzpicture}

    \begin{axis}[
        xbar,
        xlabel={Wall time (ms)},
        ylabel={Model},
        symbolic y coords={
            Pixtral5,
            Pixtral2,
            Pixtral1,
            Pixtral0,
            Llava5,
            Llava2,
            Llava1,
            Llava0,
            BLIP2, BLIP, ViT-large, ViT-base, ResNet, BERT-large, BERT-base, TinyBERT,
            },
        ytick=data,
        width=7.7cm,
        enlarge y limits=0.04,
        height=7.3cm,
        bar width=4pt,
        xmax=10000,
        xmode=log,
        error bars/x dir=both,
        error bars/x explicit,
        error bars/error bar style={black, line width=1pt}, 
        error bars/error mark options={line width=1pt, mark size=0pt},
        set layers,
    ]
        \addplot+ [
            ibm1, 
            error bars/.cd, 
            x dir=both, 
            x explicit,
        ] coordinates {
            (23,TinyBERT) +- (1.6,0)
            (267,BERT-base) +- (14.7,0)
            (911,BERT-large) +- (111,0)
            (45,ResNet) +- (179,0)
            (144,ViT-base) +- (415,0)
            (455,ViT-large) +- (1320,0)
            (862,BLIP) +- (1107,0)
            (1900,BLIP2) +- (3794,0)
            (1269,Llava0) +- (200,0)
            (2127,Llava1) +- (362,0)
            (3119,Llava2) +- (344,0)
            (6243,Llava5) +- (594,0)
            (1215,Pixtral0) +- (506,0)
            (1967,Pixtral1) +- (648,0)
            (2802,Pixtral2) +- (759,0)
            (5165,Pixtral5) +- (1452,0)
        };
    \end{axis}

\end{tikzpicture}
    \caption{Average (bar, blue) and standard deviation (line, black) wall time (in milliseconds) per sample.}
    \label{fig:wall-time}
\end{figure}
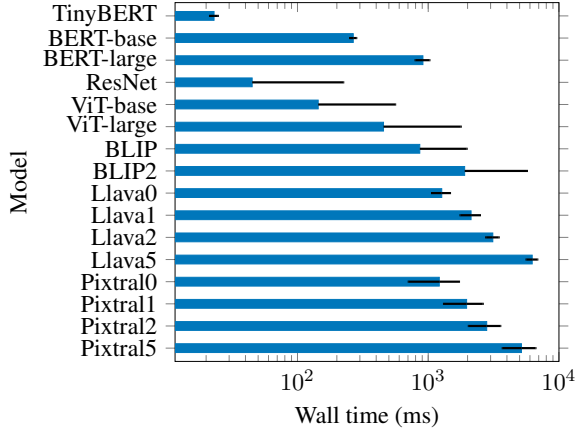

We compute the average wall time per sample and its standard deviation over 100 random samples. See Figure~\ref{fig:wall-time} for the results, including the various $n$-shot ICL models. For the FT image-based and multimodal models, the standard deviation is considerable, due to having to resize images at inference time to be fed as input to the model. Pixtral has larger standard deviations than Llava, possibly due to the larger context size in combination with KV caching.

\subsection{Prompt Error Rates}
\begin{figure}[t]
    \centering
    \begin{tikzpicture}
    \begin{axis}[
        name={ct},
        title={\textbf{CheckThat}},
        xlabel={$n$ shot},
        scaled y ticks=false, 
        yticklabel style={/pgf/number format/.cd, fixed, precision=2},
        ylabel={Error rate (\%)},
        width=7cm,
        legend to name={legend-icl-error},
        legend columns=2, 
        ymax=0.2,
        height=3cm,
        legend pos=south east,
        legend style = {
            draw=none
        },
        xtick=data,
        ymajorgrids=true,
        grid style=dashed,
        ybar,
        ymin=0,
        bar width=5pt, 
        enlarge x limits=0.2, 
        symbolic x coords={0, 1, 2, 5}, 
    ]

    \addplot[
        color=blue,
    ]
    coordinates {
(0,0.0018248175182481751)
(1,0.01824817518248175)
(2,0.0018248175182481751)
(5,0.0)
    };
    \addlegendentry{Llava, short}

    \addplot[
        color=blue,
        pattern=dots,
    ]
    coordinates {
(0,0.0)
(1,0.09854014598540146)
(2,0.025547445255474453)
(5,0.0)
    };
    \addlegendentry{Llava, long}

    \addplot[
        color=red,
    ]
    coordinates {
(0,0.0)
(1,0.014598540145985401)
(2,0.0)
(5,0.085)
    };
    \addlegendentry{Pixtral, short}

    \addplot[
        color=red,
        pattern=dots
    ]
    coordinates {
(0,0.0)
(1,0.0)
(2,0.0)
(5,0.050)
    };
    \addlegendentry{Pixtral, long}

    \end{axis}

    \begin{axis}[
        at={($(ct.south)+(0cm, -1.95cm)$)},
        name={ob},
        anchor=north,
        title={\textsc{\textbf{HintsOfTruth}}},
        xlabel={$n$ shot},
        ymax=0.2,
        scaled y ticks=false, 
        yticklabel style={/pgf/number format/.cd, fixed, precision=2},
        ylabel={Error rate (\%)},
        width=7cm,
        height=3cm,
        xtick=data,
        ymajorgrids=true,
        grid style=dashed,
        ybar,
        ymin=0,
        bar width=5pt, 
        enlarge x limits=0.2, 
        symbolic x coords={0, 1, 2, 5}, 
    ]
    \addplot[
        color=blue,
    ]
    coordinates {
(0,0.06111456391133209)
(1,0.11271911944557685)
(2,0.009987770077456177)
(5,0.0007073803348266918)
    };
        \addplot[
        color=blue,
        pattern=dots,
    ]
    coordinates {
(0,0.0009199632014719411)
(1,0.13187933143090094)
(2,0.0805136567468406)
(5,0.004280472890338361)
    };
        \addplot[
        color=red,
    ]
    coordinates {
(0,0.0012229922543823889)
(1,0.011822258459029759)
(2,0.0008153281695882593)
(5,0.006872852233676976)
    };
        \addplot[
        color=red,
        pattern=dots,
    ]
    coordinates {
(0,0.0014268242967794538)
(1,0.0040766408479412965)
(2,0.0006114961271911944)
(5,0.001661681621801263)
    };
    \end{axis}
    \node at ($(ct.east)!0.5!(ct.east)+(-2.66cm, 2.5)$) {\ref{legend-icl-error}};

\end{tikzpicture}
    \caption{Response error rates for the ICL setup.}
    \label{fig:icl-results-error}
\end{figure}
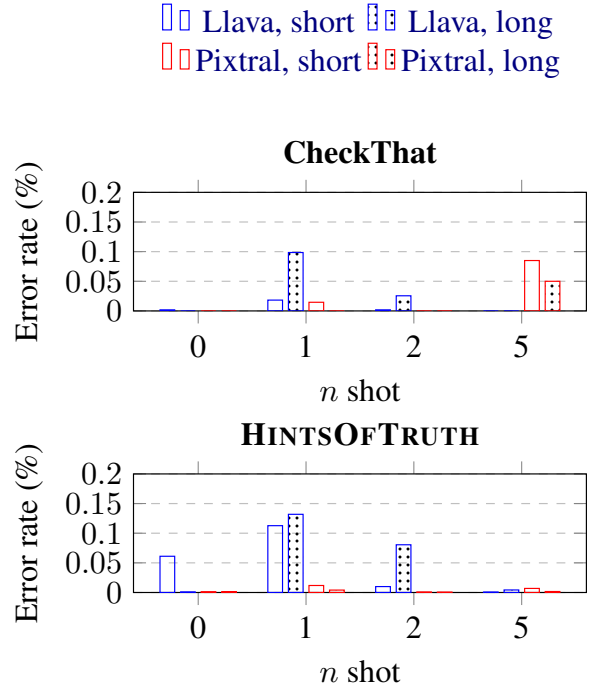

We prompt the LLMs to produce a response according to a fixed format and include some additional manual response interpretation steps to ensure we can extract the prediction from the model. However, even with this generous interpretation, the model occasionally erroneously reverts to a different response format or fails to provide a prediction label. We consider those responses as errors, and plot the error rates in Figure~\ref{fig:icl-results-error}.

The error rates are generally low, especially for the Pixtral model. For Llava, in a one-shot setup, the error rate is largest, both for CheckThat and \hiot. Qualitative observations revealed that reasons for the models failing to respond in almost all cases are due to the model refusing to answer and immediately generating an EOS token. Possibly, due to the sensitive nature of some of the claims, the samples ending up as examples in 1- and 2-shot prompts may hit a safety barrier. The low error rates highlight the robustness of the Pixtral model in adhering to the specified response format. It's discrepancy with Llava underscores the importance of model selection and prompt engineering in minimizing errors and ensuring reliable predictions.

\end{document}